\documentclass[10pt,twocolumn,letterpaper]{article}

\usepackage[pagenumbers]{cvpr} 

\usepackage[dvipsnames]{xcolor}

\definecolor{cvprblue}{rgb}{0.21,0.49,0.74}
\usepackage[pagebackref,breaklinks,colorlinks,citecolor=cvprblue]{hyperref}
\usepackage{multirow}
\usepackage{amssymb}
\usepackage{xcolor}
\usepackage{pifont}
\usepackage{tikz}
\usetikzlibrary{shapes}
\newcommand*\colourcheck[1]{%
  \expandafter\newcommand\csname #1check\endcsname{\textcolor{#1}{\ding{52}}}%
}
\colourcheck{green}
\colourcheck{black}

\newcommand*\colourx[1]{%
  \expandafter\newcommand\csname #1x\endcsname{\textcolor{#1}{\ding{55}}}%
}
\colourx{red}

\newcommand{\mytriangle}[1]{\tikz{\node[draw=#1,fill=#1,isosceles
triangle,isosceles triangle stretches,shape border rotate=90,minimum
width=0.2cm,minimum height=0.2cm,inner sep=0pt] at (0,0) {};}}

\usepackage[linesnumbered,ruled,vlined]{algorithm2e}
\usepackage {algpseudocode}
\usepackage{algorithmicx}
\usepackage{algcompatible}

\usepackage[accsupp]{axessibility} 

\def\paperID{5394} 
\def\confName{CVPR}
\def\confYear{2024}

\title{WWW: A Unified Framework for Explaining \textbf{W}hat, \textbf{W}here and \textbf{W}hy of Neural Networks by Interpretation of Neuron Concepts}

\author{Yong Hyun Ahn, 
Hyeon Bae Kim, 
Seong Tae Kim\thanks{Dr. S.T. Kim is corresponding author.}
\\
Kyung Hee University, Republic of Korea \\
{\tt\small \{yhahn, hyeonbae.kim, st.kim\}@khu.ac.kr}\\
}

\begin{document}
\maketitle

\begin{abstract} 
Recent advancements in neural networks have showcased their remarkable capabilities across various domains. 
Despite these successes, the ``black box” problem still remains. 
To address this, we propose a novel framework, WWW, that offers the `what', `where', and `why' of the neural network decisions in human-understandable terms. Specifically, WWW utilizes adaptive selection for concept discovery, employing adaptive cosine similarity and thresholding techniques to effectively explain `what'.
To address the `where' and `why', we proposed a novel combination of neuron activation maps (NAMs) with Shapley values, generating localized concept maps and heatmaps for individual inputs. 
Furthermore, WWW introduces a method for predicting uncertainty, leveraging heatmap similarities to estimate the prediction's reliability. 
Experimental evaluations of WWW demonstrate superior performance in both quantitative and qualitative metrics, outperforming existing methods in interpretability. 
WWW provides a unified solution for explaining `what', `where', and `why', introducing a method for localized explanations from global interpretations and offering a plug-and-play solution adaptable to various architectures. 
Code is available at: \href{https://github.com/ailab-kyunghee/WWW}{https://github.com/ailab-kyunghee/WWW}.
\end{abstract}    
\section{Introduction}
\label{sec:intro}

Neural networks have demonstrated impressive performance in various fields in recent years.
Despite these successes, their widespread adoption in more diverse areas is slowed by several challenges.
A fundamental issue is the ``black box" problem, referring to the often hidden and unclear decision-making processes of neural network models. 
This lack of clarity raises concerns about the stability and reliability of these models, leading to a growing consensus that artificial intelligence should be reliable, robust, and safe~\cite{doshi2017towards, jacovi2021formalizing}.
As a response to this need for trustworthy AI, there has been an emergence of laws and regulations~\cite{kaminski2021right,kop2021eu} that require neural networks to base their decisions on principles that are understandable to humans.

\begin{table}[t]
\caption{\small \textbf{Illustration of What, Where, and Why of recent concept-based neural network interpretation methods.} 
Green, yellow, and red marks illustrate that the method is able to interpret well, is partially interpretable, and has limitations for interpretation, respectively.
Recent methods are able to interpret one or two `w's but have limitations regarding interpreting three `w's at once.
}
\label{tab:part-abl}
\scalebox{0.92}{
\centering
    \begin{tabular}{lccc}
    \toprule
    \textbf{Method} & \textbf{What} & \textbf{Where} & \textbf{Why} \\ 
    \midrule
    CLIP-Dissect~\cite{oikarinen2023clipdissect} (ICLR'23) & \greencheck &  \redx  & \redx \\ 
    FALCON~\cite{kalibhat2023identifying} (ICML'23) &\greencheck  &\redx &  \mytriangle{yellow}  \\
    CRAFT~\cite{fel2023craft} (CVPR'23)& \redx & \greencheck & \mytriangle{yellow} \\
    \textbf{WWW (Ours)} &\greencheck  &\greencheck & \greencheck  \\
    \bottomrule
    \end{tabular}
    }
\end{table}

According to Doshi-Velez~\textit{et al.}~\cite{doshi2017towards}, interpretability is defined as the capability to provide explanations in terms that are understandable to humans. 
Zhang~\textit{et al.}~\cite{zhang2021survey} expand on this definition, emphasizing that interpretability involves providing explanations in understandable terms to humans. 
This requirement for explanations in human-understandable terms is a consistent theme in the literature on interpretable methods.
Furthermore, this concept aligns with a fundamental principle in journalism and problem-solving: the importance of clear and understandable communication. 
This principle is often encapsulated in the five Ws (Who, What, When, Where, Why)~\cite{sloan2010aristotle}.
Building on this, we propose that practical explanations or interpretations should include three critical elements: `what' (the nature of the decision or outcome), `where' (the context or specific region of the input that affects the decision-making process), and `why' (the reasoning or factors behind a decision). 
This approach aims to make the interpretations more accessible and relevant to human understanding.

From this point of view, former works can be re-grouped into several groups which those works aim to solve.
For example, a series of works in feature attribution~\cite{selvaraju2017grad,fong2017interpretable,zhang2021fine} can be classified as works mainly focusing on `where'.
These methods explain model decisions by identifying highly contributed input regions.
However, identifying highly related input regions primarily focuses on `where' only, not enough explanation for `what' and `why'.

On the other hand, concept-based explanations mainly focus on `why'.
Concept-based explanations~\cite{kim2018interpretability,fong2017interpretable, fel2023holistic,fel2023craft} explain model decisions by decomposing model responses into smaller units called concepts.
Kim~\textit{et al.} proposed TCAV~\cite{kim2018interpretability}, understanding and interpreting model decisions into a form of concept activation vectors. 
Recently, Fel~\textit{et al.} proposed CRAFT~\cite{fel2023craft}, which utilizes non-negative matrix factorization for identifying concept vectors.
The works mentioned above mainly decompose activations into the form of vectors, which gives a great advantage for finding out `why' but often fails to match or annotate the human-understandable term (i.e., name) that is needed for `what'.

Other streams of work, called Neuron-concept association, aim to name what each neuron (e.g. convolution filters, layer outputs) represents, which answers `what' the model cares about.
Bau \textit{et al.} proposed Network Dissection~\cite{bau2017network}, which identifies each neuron's representing concept from Broaden dataset~\cite{bau2017network}.

Recently, CLIP-Dissect~\cite{oikarinen2023clipdissect} and FALCON~\cite{kalibhat2023identifying} have shown that representing the concept of neurons can be matched automatically by using a pre-trained CLIP model~\cite{radford2021clip}. 
The aforementioned methods are suitable for explaining `what' but partially explain or have limitations for explaining `where' and `why'.
Furthermore, these methods generate global explanations focusing on each neuron's role in the network, not the local explanations for each sample input.

To address these issues, we propose a novel framework that can explain `\textbf{W}hat', `\textbf{W}here', and `\textbf{W}hy' (WWW) at once.
WWW introduces adaptive selection for discovering each neuron's concept and interpreting each neuron to explain `what'.
By leveraging adaptive cosine similarity (ACS) and adaptive selection techniques, we achieve advanced performance compared to competitive methods.
Moreover, we combined neuron activation map (NAM) and Shapley value~\cite{shapley1997value,khakzar2021neural,ahn2023line} to generate class and sample concept maps and heatmaps to explain `where' and `why'. 
We also conducted various objective evaluations to assess the performance of the proposed method.
In the experiments, WWW achieves better results in both quantitative and qualitative evaluations.
Our key contributions are summarized as follows:
\begin{itemize} 
    \item 
    We introduce a novel and effective way to generate high-quality explanations that explain `what', `where', and `why' at once. Due to the powerful performance of the adaptive selection for concept discovery, WWW is able to achieve higher quantitative and better qualitative results in various metrics. 
    \item 
    We introduce a way to generate localized explanations from global neural network interpretation. By the novel combination of Shapley value and neuron activation maps, WWW is able to generate localized explanations with concept annotations for sample input.
    \item
    WWW can be attached to various target models with different architectures, from conventional convolution neural networks to the recent attention-based Vision transformers, in a plug-and-play manner. 
\end{itemize}

\section{Related Works}
\label{sec:related}
\subsection{Neuron-Concept Assosiation}
Bau~\textit{et al.} introduced Network Dissection \cite{bau2017network}, using the Broden dataset to identify which concepts individual neurons in a network represent. They use overlap between segmentation masks and feature maps to annotate concepts for neurons. 
Fong and Vedaldi expanded on this with Net2Vec~\cite{fong2018net2vec}, which looks at individual neurons and their combinations.
Mu and Andreas further extended these ideas with Compositional Explanation~\cite{mu2020compositional}, aiming to generate more complex and detailed explanations.
These methods primarily focused on understanding `what' a neuron represents. 
However, the aforementioned methods fell short in explaining the `where' and `why' of neuron representations. 
Additionally, they relied heavily on image-concept-matched datasets like Broden, which are often costly and hard to collect due to the need for pixel-wise labels.
To address these limitations, recent approaches like HINT~\cite{wang2022hint} and MILAN~\cite{MILAN} have been developed.
These methods train concept classifiers or models that reduce the dependency on image-concept-matched datasets. 
Moreover, approaches like CLIP-Dissect~\cite{oikarinen2023clipdissect} and FALCON~\cite{kalibhat2023identifying} leverage the pre-trained CLIP model to use separate sets of image datasets and concept datasets.
Despite these advancements in understanding `what' neurons represent, there remains a gap in fully explaining the `where' and `why' of the neuron representations. 

\begin{figure*}[t]
    \centering
    \includegraphics[width=0.92\textwidth]{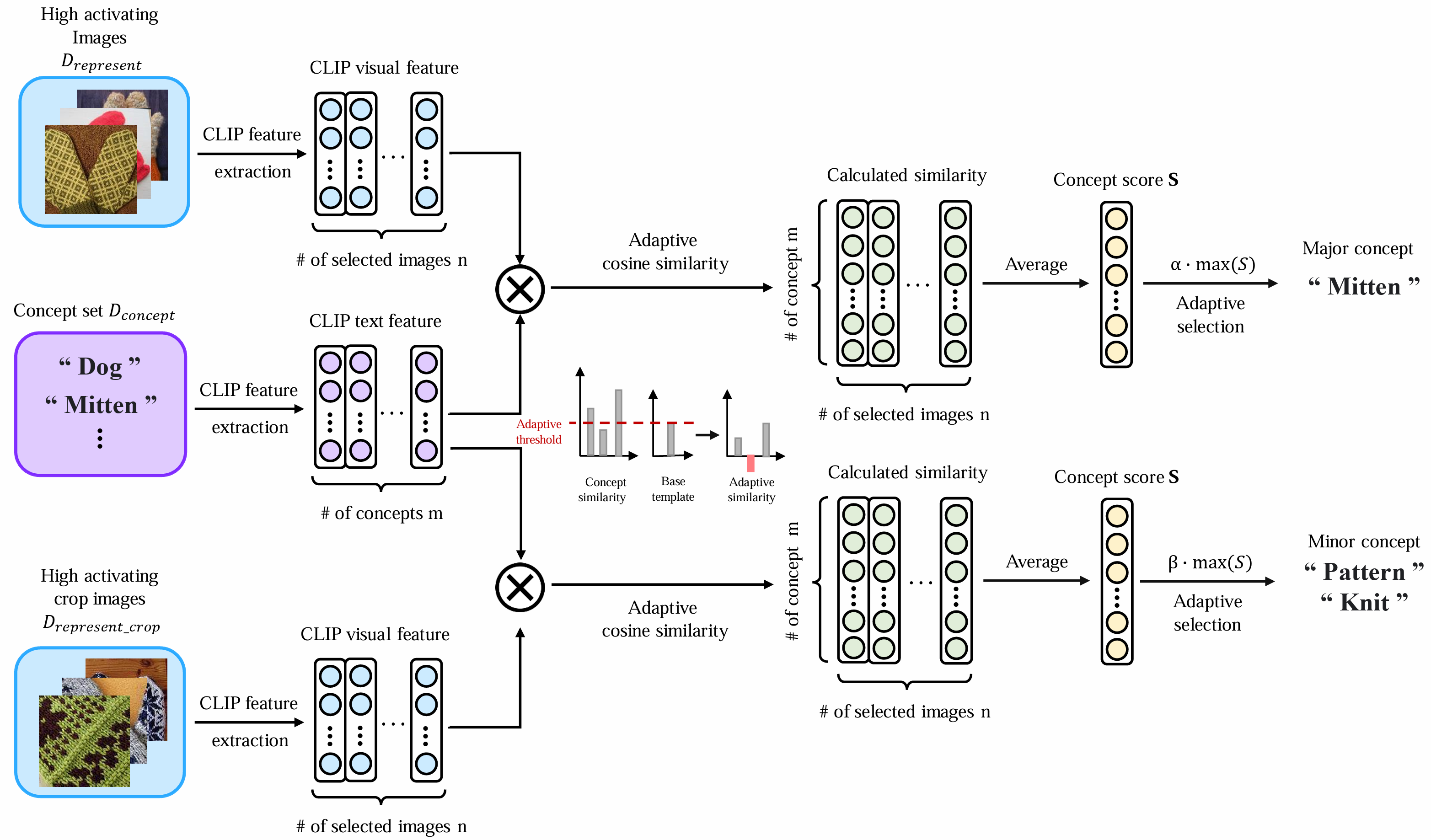}
    \caption{\small\textbf{Overall flow of Concept Discovery module identifying concepts for a single neuron.}
    We first calculate the cosine similarity of CLIP features between images and concepts with the template from the selected high-activating images. 
    Then, we subtract the cosine similarity of CLIP features between images and the base template by only considering the similarity between the concept and image. 
    From calculated adaptive cosine similarity (ACS), we generate concept score $S$ by the average similarity of images. 
    Note that concept score $S=\{s_1,s_2,...,s_m\}$ are a group of scores, not a single scalar. 
    From the calculated concept scores $S$, we select major concepts by adaptive selection.
    We also discover minor concepts using the same process but with crop images.
    }
    \label{fig:concept_extract} 
\end{figure*}

\subsection{Vector-based Explanation}
The field of interpretability has significantly advanced, particularly in understanding and interpreting model decisions using Concept Activation Vectors (CAVs). 
Kim~\textit{et al.} introduced TCAV~\cite{kim2018interpretability}, understanding and interpreting model decisions into the form of CAVs.
Recently, CRAFT~\cite{fel2023craft} was introduced, leveraging non-negative matrix factorization to identify CAVs and localize the most relevant input regions for each CAV.
Also, Achtibat~\textit{et al.}~\cite{achtibat2023attribution} and Fel~\textit{et al.}~\cite{fel2023holistic} proposed methods to interpret each CAV's role in decision-making.
Despite these developments, a key challenge remains: translating CAVs into human-understandable terms.

\section{Method}
\subsection{Method Overview}
WWW consists of three modules: Concept discovery, Localization, and Reasoning.
The concept discovery module identifies each neuron's concept, which explains part `what'.
It selects neuron concepts from the concept set ($D_{concept}$) by leveraging adaptive cosine similarity and adaptive selection.
The localization module is to identify highly contributed input regions of the test sample, which concept is present at `where'.
Also, the combination of the neuron activation map leveraging Shapley value helps identify `where' the concept is and tells the `why' of the individual predictions.
The reasoning module identifies important neurons of the test sample and predicted class.
By comparing the differences between the sample and class explanations, users can understand the `why' of the model prediction and, even more, whether the prediction is reliable or not.

\begin{figure*}[t]
    \centering
    \includegraphics[width=\textwidth]{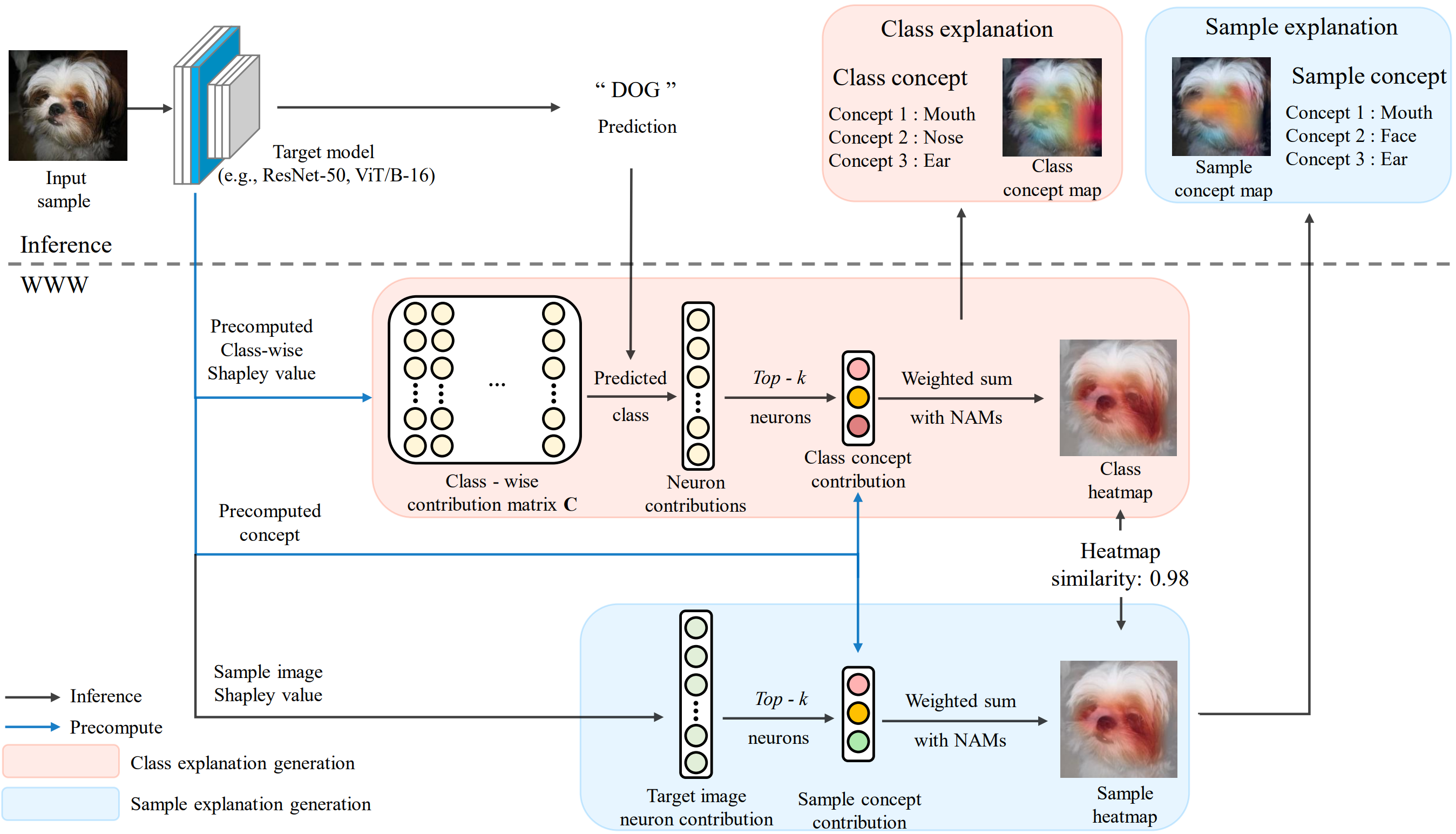}
    \caption{\small\textbf{Illustration of overall test time flow of WWW.}
    In the test time (i.e., inference), class explanation selects important neurons with pre-computed Shapley value of the predicted class. On the other hand, the sample explanation selects important neurons with a Shapley value for the input sample. With selected important neurons, pre-computed concepts are annotated. After the concept annotation, a class heatmap is generated with the pre-computed Shapley value of the predicted class. On the other hand, a sample heatmap is generated with the Shapley value of the sample input.
    }
    \label{fig:concept} 
\end{figure*}

\subsection{Concept Discovery Module}
\label{sec:CDM}
The concept discovery module aims to identify proper major and minor concepts from $D_{concept}$ that match an example-based representation of each neuron.
Let target model $f$ d, and let $(l,i)$ as $i$-th neuron in layer $l$ of the target model.
From the images in probing dataset $D_{probe}$, we select high activating images for neuron $(l,i)$ as
$D^{(l,i)}_{rep}$.
With selected images in $D^{(l,i)}_{rep}$ and concepts in $D_{concept}$, we calculate CLIP visual feature of $D^{(l,i)}_{rep}$ as $V^{(l,i)} = \left[v^{(l,i)}_{1}, v^{(l,i)}_{2}, \cdots,v^{(l,i)}_{n} \right]$ and CLIP text feature of $D_{concept}$ as $T = \left[t_{1}, t_{2}, \cdots,t_{m} \right]$.
$n$ denotes number of example images for neuron $(l,i)$ and $m$ stands for the number of concept in concept set $D_{concept}$.

We calculate concept score $s^{(l,i)}$ by calculating adaptive cosine similarity (ACS), which allows us to reduce the effect of the base template and only consider the similarity between the image and the concept itself. 
Concept score $s^{(l,i)}_{j}$ for $j$-th concept $t_{j}$ of neuron $(l,i)$ is calculated as follows:
\begin{equation}\label{equ:Adaptive_sim}
    s^{(l,i)}_{j} = \frac{1}{n}\sum^{n}_{o = 1} \left\{cos(v^{(l,i)}_{o}, t_{j}) - cos(v^{(l,i)}_{o},t_{tem}) \right\},
\end{equation}
where $1\leq j \leq m$, $t_{tem}$ denotes CLIP text feature of base templete(e.g. `a photo of.') and $cos(x,y)$ denotes cosine similarity between $x$ and $y$.

From the calculated concept scores $s^{(l,i)} = \left[s^{(l,i)}_{1},\cdots, s^{(l,i)}_{m} \right]$, we select concepts where $s^{(l,i)}_{j} >  \delta^{(l,i)}$.

Adaptive selection threshold $\delta^{(l,i)}$ for neuron $(l,i)$ is calculated as follows:
\begin{equation}\label{equ:Adaptive_select}
    \delta^{(l,i)} = \alpha \times max(s^{(l,i)}),\\
\end{equation}
where $\alpha$ denotes concept sensitivity for major and minor concepts.
For discovering minor concepts, we select $D^{(l,i)}_{rep}$ with cropped images of probing dataset $D_{probe}$.

\subsection{Localization Module}
\label{Sec:LM}
The localization module aims to identify highly contributed input regions of each concept and generate a concept map and concept heatmap.
We select important neurons with Taylor approximation of Shapley value introduced in \cite{khakzar2021neural} for generating a concept region map. 
For input image sample $x$, Neuron contribution $w^{(l,i)}$ of neuron $(l,i)$ is calculated as follows:
\begin{equation}\label{equ:taylor}
    w^{(l,i)}(x) = \left| f(x) - f(x;a^{(l,i)}\leftarrow 0)\right| = \left| a^{(l,i)}\nabla_{a^{(l,i)}}f(x) \right|.
\end{equation}
Where $a^{(l,i)}$ denotes activation of neuron $(l,i)$.
After calculating each layer's Neuron contribution, we rank neurons by calculated contribution (i.e., Shapley value) and select the top-$k$ important neurons for the sample.
The concept heatmap $M$ is calculated with the weighted sum of important neurons' neuron activation maps (NAMs).
Concept heatmap $M^{l}(x)$ of important neurons is calculated as follows:
\begin{equation}\label{equ:Adaptive_select}
    M^{l}(x) = \sum^{u}{w^{(l,u)}(x)A^{(l,u)}(x)}
\end{equation}
where $u$ denotes the index of important neurons, $w$ denotes the neuron contribution, and $A$ denotes the neuron activation map of the neuron.
Note that $M^{l}$ shows only related regions of a single or combination of important neurons, not the whole network.

\subsection{Reasoning Module}
\label{Sec:RM}
The reasoning module is designed to help users understand the `why' of the model output.
This not only explains the result but also includes `why' this prediction is reliable or not.
By leveraging the class-wise Shapley value introduced in \cite{ahn2023line}, we can understand each neuron's class-wise contribution and identify important neurons for each class.
From Ahn \textit{et al.}~\cite{ahn2023line}, the class-wise contribution of a neuron can be calculated by the average contribution of a neuron in class samples.
From the class-wise contribution, we rank top-$k$ important neurons for each class.
We can also generate a class-wise concept region map and concept heatmap with the index of important neurons.
Class-wise maps can be used as a guideline for understanding the general case of the prediction.
By comparing Class-wise maps and sample maps, users can identify which concept and region of the sample differs from general cases.
This naturally helps users understand the `why' of the model output and also `why' the results are reliable or not.
Figure~\ref{fig:concept} for further understanding the WWW flow for generating class and sample explanations. Figure~\ref{fig:out_success} shows the generated explanation example.

\subsection{Overall flow and Output of WWW}
\label{sec:overall_flow}
As described above, WWW has three main modules explaining three `w' of `what', `where', and `why', respectively.
In Section \ref{sec:overall_flow}, we are going to follow the overall flow of our method by time sequence.
Before starting inference, WWW needs pre-computing for discovering major and minor concepts with the concept discovery module (Sec~\ref{sec:CDM}) and also a class-wise contribution for class-wise analysis in the Reasoning Module (Sec~\ref{Sec:RM}).
After pre-computing concepts and contributions, WWW is ready to generate test sample explanations.
Figure \ref{fig:concept} shows the overall flow of explaining the generation of WWW in the inference time.
In the test time (i.e., inference), WWW generates a class-wise explanation of the predicted class and sample explanation into two tracks.
WWW leverages the predicted class and class-wise contribution matrix to generate a class-wise explanation to select the most critical neurons for the predicted class.
After identifying important neurons, WWW leverages the localization module (Sec~\ref{Sec:LM}) to generate a class concept region map and class concept combination heatmap.
On the other hand, for the sample explanation, WWW calculates the Taylor-approximated Shapley value of the sample for the predicted class and identifies critical neurons for the prediction of the sample.
After identifying important neurons for the sample, WWW leverages the sample Shapley value to generate the sample concept region map and sample concept combination heat map.

\section{Experiment}
Comprehensive experiments have been conducted to evaluate our method.
In section~\ref{sec:eval_CDM}, we evaluate the performance of the concept discovery module both qualitatively (Sec.~\ref{eval_cdm_qual}) and quantitatively (Sec.~\ref{eval_cdm_quan}).
Section~\ref{sec:abl} is an ablation study for the concept discovery module.
In Section~\ref{sec:Discussion}, we analyze generated explanations of both correct and wrong predictions with examples.
\subsection{Performance Evaluation for Concept Module}
\label{sec:eval_CDM}
In this experiment, we evaluate the concept matching performance of the concept discovery module with four other baselines, Network Dissection~\cite{bau2017network}, MILAN~\cite{MILAN}, CLIP-Dissect~\cite{oikarinen2023clipdissect} and FALCON~\cite{kalibhat2023identifying}.
We evaluate the performance of methods on the various models (e.g., ResNet-18~\cite{resnet}, ResNet-50~\cite{resnet}, and ViT-B/16~\cite{vit}), various probing datasets (e.g., Imagenet, Places365) and various concept sets (e.g., Wordnet nouns, and labels of Places365, Broaden, and ImageNet)
\begin{figure}[t!]
    \centering
    \includegraphics[width=0.48\textwidth]{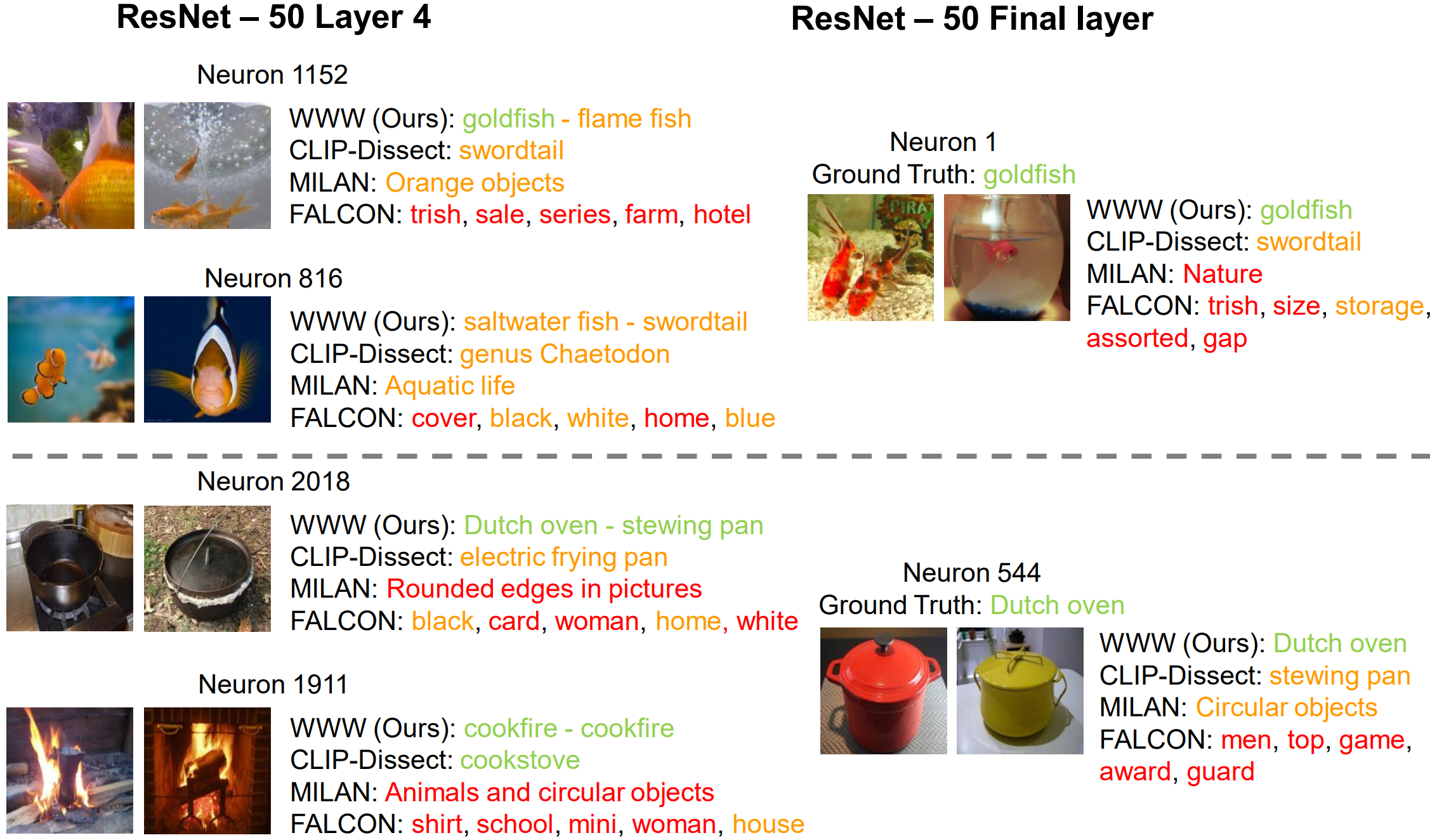}
    \caption{\small\textbf{Qualitative comparison of WWW with other baselines.} We compared WWW with three competitive baselines (CLIP-Dissect~\cite{oikarinen2023clipdissect}, MILAN~\cite{MILAN}, FALCON~\cite{kalibhat2023identifying}) in two final layer neurons and four penultimate layer (i.e., layer 4) neurons with each neuron's highly activating images. layer-4 neurons are top-$2$ important neurons of the final layer class. We have colored the descriptions green if they match the images, yellow if they match but are too generic or similar, and red if they do not match.
    }
    \label{fig:qual_CDM} 
\end{figure}
\subsubsection{Qualitative Results}
\label{eval_cdm_qual}
\begin{table*}[t!]
\caption{\textbf{Quantitative comparison on final layer concept matching performance of ResNet-50 trained on ImageNet.} We compared predicted neuron concepts with ground truth labels of ImageNet. We used the Imagenet-1k validation set for $D_{probe}$.
\textbf{Bold} numbers represent the best scores between the same settings. The average score and standard errors of the $1000$ final layer neurons are reported.
}
\centering
\scalebox{0.90}{
\begin{tabular}{@{}l|ll|cccc@{}}
\toprule
Method & $D_{probe}$ & $D_{concept}$ & CLIP cos & mpnet cos & F1-score & Hit Rate\\ \midrule
Network Dissection~\cite{bau2017network} & Broden & Broden(1.2k) & $0.7229\pm{0.003}$ & $0.2989\pm{0.005}$ & $0.0010\pm{0.001}$ & 0.001\\ 
MILAN(b)~\cite{MILAN} & ImageNet val & - & $0.7300\pm{0.003}$ & $0.2485\pm{0.005}$ & $0.0005\pm{0.000}$ & 0.001 \\
FALCON~\cite{kalibhat2023identifying}  & ImageNet val & LAION-400m & $0.7065\pm{0.003}$ & $0.1790\pm{0.001}$ & $0.0002\pm{0.000}$ & 0.001 \\
\midrule
\multirow{3}{*}{CLIP-Dissect~\cite{oikarinen2023clipdissect}}
  & ImageNet val & ImageNet (1k) & $\mathbf{0.9340\pm{0.003}}$ & $\mathbf{0.8376\pm{0.006}}$ & $0.7286\pm{0.009}$ & 0.933\\
  & ImageNet val & Broden (1.2k) & $0.7369\pm{0.003}$ & $0.3432\pm{0.004}$ & $0.0328\pm{0.003}$ & \textbf{0.108}\\
  & ImageNet val & Wordnet (80k) & $0.8689\pm{0.004}$ & $0.6846\pm{0.008}$ & $0.3647\pm{0.014}$ & 0.456\\    
\midrule
\multirow{3}{*}{WWW (Ours)}
  & ImageNet val & ImageNet (1k) & $0.9325\pm{0.003}$ & $0.8327\pm{0.006}$ & $\mathbf{0.7719\pm{0.009}}$ & \textbf{0.955}\\
  & ImageNet val & Broden (1.2k) & $\mathbf{0.7758\pm{0.004}}$ & $\mathbf{0.4414\pm{0.007}}$ & $\mathbf{0.0645\pm{0.007}}$ & 0.091\\
  & ImageNet val & Wordnet (80k) & $\mathbf{0.8858\pm{0.003}}$ & $\mathbf{0.6945\pm{0.008}}$ & $\mathbf{0.4197\pm{0.012}}$ & \textbf{0.645}\\
\bottomrule

\end{tabular}
}
\label{tab:eval_res50}
\end{table*}

\begin{table*}[t!]
\caption{\textbf{Quantitative comparison on final layer concept matching performance of ViT-B/16 trained on ImageNet.} We compared predicted neuron concepts with ground truth labels of ImageNet. We used the Imagenet-1k validation set for $D_{probe}$.
\textbf{Bold} numbers represent the best scores between the same settings. The average score and standard errors of the $1000$ final layer neurons are reported.
}
\centering
\scalebox{0.90}{
\begin{tabular}{@{}l|ll|cccc@{}}
\toprule
Method & $D_{probe}$ & $D_{concept}$ & CLIP cos & mpnet cos & F1-score & Hit Rate\\
\midrule
\multirow{3}{*}{CLIP-Dissect~\cite{oikarinen2023clipdissect}}
  & ImageNet val & ImageNet (1k) & $\mathbf{0.9337\pm{0.003}}$ & $\mathbf{0.8375\pm{0.006}}$ & $0.7289\pm{0.009}$ & 0.933\\
  & ImageNet val & Broden (1.2k) & $0.7365\pm{0.003}$ & $0.3416\pm{0.004}$ & $0.0319\pm{0.003}$ & \textbf{0.106}\\
  & ImageNet val & Wordnet (80k) & $0.8700\pm{0.004}$ & $0.6886\pm{0.008}$ & $0.3679\pm{0.014}$ & 0.460\\
\midrule
\multirow{3}{*}{WWW (Ours)}
  & ImageNet val & ImageNet (1k) & $0.9331\pm{0.003}$ & $0.8347\pm{0.006}$ & $\mathbf{0.7718\pm{0.009}}$ & \textbf{0.955}\\
  & ImageNet val & Broden (1.2k)   & $\mathbf{0.7754\pm{0.004}}$ & $\mathbf{0.4407\pm{0.007}}$ & $\mathbf{0.0651\pm{0.007}}$ & 0.091\\
  & ImageNet val & Wordnet (80k)  & $\mathbf{0.8857\pm{0.003}}$ & $\mathbf{0.6970\pm{0.008}}$ & $\mathbf{0.4166\pm{0.012}}$ & \textbf{0.634}\\
\bottomrule
\end{tabular}
}
\label{tab:eval_vit}
\end{table*}

\noindent\textbf{Settings.}
We compared WWW with the three most comparable methods (CLIP-Dissect~\cite{oikarinen2023clipdissect}, MILAN~\cite{MILAN}, FALCON~\cite{kalibhat2023identifying}) in the penultimate layer and final layer of the model.
We do not compare with Network Dissection~\cite{bau2017network} due to the limitation that probe image data $D_{probe}$ and concept set $D_{concept}$ is fixed to Broaden.
We used a ResNet-50~\cite{resnet} model pre-trained in the ImageNet-1k~\cite{imagenet} dataset.
For probe image data $D_{probe}$, we used the ImageNet-1k validation set, and we extracted all nouns in Wordnet\cite{wordnet} (about 80k) dataset for a concept set $D_{concept}$.

\noindent\textbf{Results.}
Figure \ref{fig:qual_CDM} shows examples of descriptions for hidden neurons in the penultimate and final layers.
Neurons in the penultimate layer are top-$2$ important neurons of the final layer neuron's ground truth label class.
We observed that WWW not only interpreted each neuron well but also showed robust interpretation that the most important neuron of the class in the penultimate layer represents the same major concept as the final layer neuron.
\subsubsection{Quantitative Results}
\label{eval_cdm_quan}
In this section, we compare the performance of our methods with baselines. 
As introduced in~\cite{oikarinen2023clipdissect}, we evaluate final layer neuron concepts with the class labels with various metrics.
By comparing generated explanations with the class labels, we can objectively evaluate the quality of the generated neuron labels with what each neuron is trained to represent.

\noindent\textbf{Metrics.} CLIP cos and mpnet cos are measured as cosine similarities between the encoded feature of the class label and selected concepts with CLIP~\cite{radford2021clip} model and mpnet~\cite{song2020mpnet} model, respectively.
F1-score is measured to evaluate the discovered concept's balance of exactness and flexibility.
Also, the hit rate is calculated as a rate of selected concepts that exactly match the class label.

\noindent\textbf{Quantitative comparison on ResNet-50.}
Table \ref{tab:eval_res50} compares WWW with Network Dissection~\cite{bau2017network}, MILAN(b)~\cite{MILAN}, FALCON~\cite{kalibhat2023identifying}, and CLIP-Dissect~\cite{oikarinen2023clipdissect}.
We evaluate the final layer neuron concepts of ResNet-50 with the class labels of ImageNet-1k.
In table \ref{tab:eval_res50}, WWW showed better performance as the concept set $D_{concept}$ gets larger and outperformed all other baselines when concept set $D_{concept}$ is set to Broden and Wordnet nouns.
In the results of Wordnet nouns ($D_{concept}$), comparison between WWW and CLIP-Dissect~\cite{oikarinen2023clipdissect} are statistically significant across all metrics ($p < 0.05$).

\noindent\textbf{Quantitative comparison on ViT-B/16.}
We compared predicted concepts of the final layer to ground truth labels of ViT-B/16 pre-trained on ImageNet.
In Table~\ref{tab:eval_vit}, WWW showed better performance as the concept set $D_{concept}$ gets larger and outperformed other baselines when concept set $D_{concept}$ is set to Broden and Wordnet nouns.
\begin{table*}[t!]
\caption{\textbf{Quantitative comparison on final layer concept matching performance of ResNet-18 trained on Places365.} We compared predicted neuron concepts with ground truth labels of Places365. We used the Places365 test set for $D_{probe}$.
\textbf{Bold} numbers represent the best scores between the same settings. The average score and standard errors of the $365$ final layer neurons are reported.
}
\centering
\scalebox{0.90}{
\begin{tabular}{@{}l|ll|cccc@{}}
\toprule
Method & $D_{probe}$ & $D_{concept}$ & CLIP cos & mpnet cos & F1-score & Hit Rate \\
\midrule
\multirow{3}{*}{CLIP-Dissect~\cite{oikarinen2023clipdissect}}
  & Places365 test & Places365 (0.4k) & $\mathbf{0.9562\pm{0.004}}$ & $\mathbf{0.8687\pm{0.012}}$ & $\mathbf{0.7233\pm{0.023}}$ & \textbf{0.723}\\
  & Places365 test & Broden (1.2k)    & $0.8304\pm{0.003}$ & $0.4678\pm{0.007}$ & $0.1096\pm{0.008}$ & \textbf{0.329}\\
  & Places365 test & Wordnet (80k)       & $0.8378\pm{0.006}$ & $\mathbf{0.5107\pm{0.014}}$ & $0.1041\pm{0.016}$ & 0.104\\
\midrule
\multirow{3}{*}{WWW (Ours)}
  & Places365 test & Places365 (0.4k) & $0.9402\pm{0.004}$ & $0.8204\pm{0.013}$ & $0.6361\pm{0.024}$ & 0.660\\
  & Places365 test & Broden (1.2k) & $\mathbf{0.8925\pm{0.005}}$ & $\mathbf{0.6415\pm{0.013}}$ & $\mathbf{0.2242\pm{0.021}}$ & 0.255\\
  & Places365 test & Wordnet (80k) & $\mathbf{0.8492\pm{0.004}}$ & $0.5000\pm{0.013}$ & $\mathbf{0.1106\pm{0.015}}$ & \textbf{0.142}\\
\bottomrule
\end{tabular}
}
\label{tab:eval_plac}
\end{table*}

\noindent\textbf{Quantitative comparison on ResNet-18 pre-trained in Places365.}
We compared predicted labels to ground truth labels in final layer neurons of ResNet-18 pre-trained on Places365.
In table~\ref{tab:eval_plac}, WWW showed better performance as the concept set $D_{concept}$ gets larger and outperformed other baselines when concept set $D_{concept}$ is set to Broden and Wordnet nouns.
\subsection{Ablation study}
\label{sec:abl}

\begin{table}[t!]
\caption{\textbf{Ablation study on the use of the template and ACS.} `\blackcheck' in the Template represents WWW using a template (e.g., `a photo of \textit{word}') `\blackcheck' in the ACS represents WWW using adaptive cosine similarity.  
}
\centering
\scalebox{0.72}{
\begin{tabular}{cc|ccc}
\toprule
Template & ACS & CLIP cos & mpnet cos & F1-score\\
\midrule
&  & $0.8499\pm{0.003}$ & $0.6123\pm{0.007}$ & $0.3265\pm{0.009}$ \\
\blackcheck &  & $0.8547\pm{0.003}$ & $0.6075\pm{0.007}$ & $0.3361\pm{0.009}$ \\
\blackcheck & \blackcheck & $\mathbf{0.8858\pm{0.003}}$ & $\mathbf{0.6945\pm{0.008}}$ & $\mathbf{0.4197\pm{0.012}}$  \\
\bottomrule
\end{tabular}
}
\label{tab:abl_acs}
\end{table}

\begin{figure}[t]
    \centering
    \includegraphics[width=0.48\textwidth]{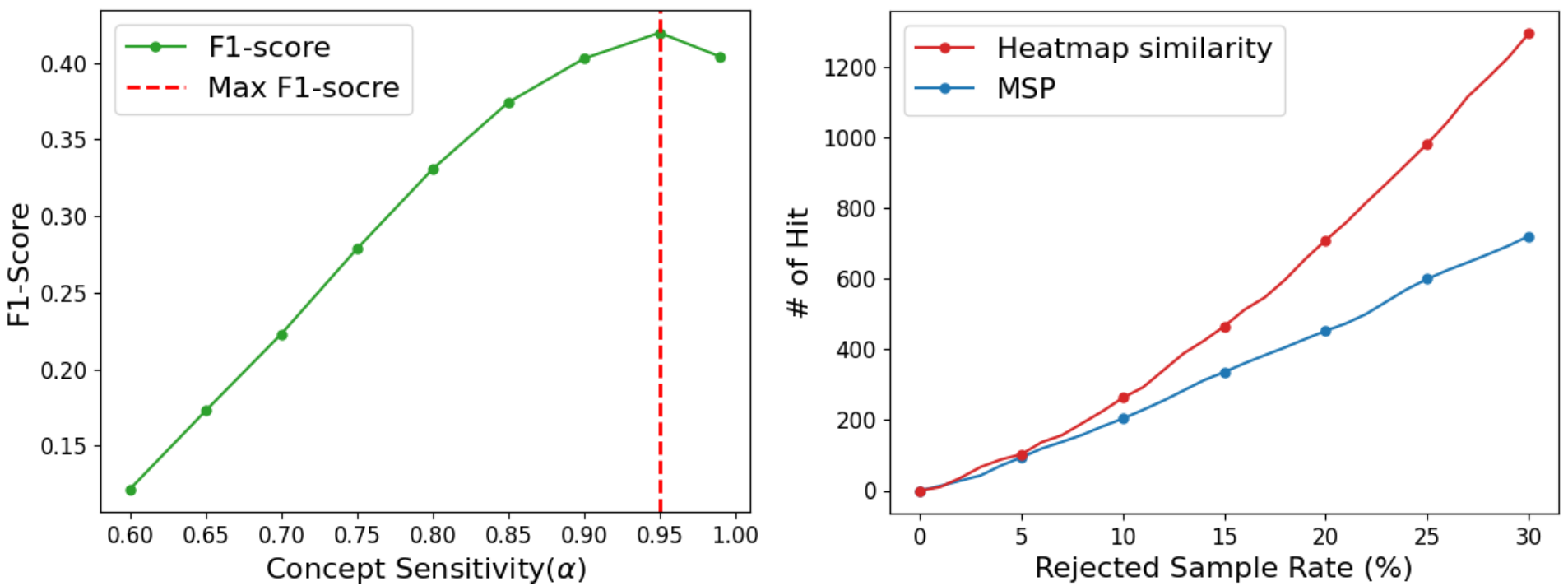}
    \caption{\small\textbf{Ablation of concept sensitivity and heatmap similarity feasibility result.} Left figure illustrates the F1 score with respect to Concept Sensitivity ($\alpha$) changes. Concept sensitivity ($\alpha$) that maximizes the F1-score is illustrated as the red line. The right figure illustrates the rejection test result of heatmap similarity and maximum softmax probability (MSP). \# of hit denotes the number of correctly detected samples as a misprediction.
    }
    \label{fig:graph} 
\end{figure}

\noindent\textbf{Evaluation on the effect of leveraging base template and ACS.}
Table~\ref{tab:abl_acs} shows an ablation study over various parts used in WWW.
Without using a base template, WWW showed the lowest performance overall.
With the use of a base template, WWW shows slightly increased performance in CLIP cos and F1-score but not much advance overall.
However, performance was significantly increased, and the highest scores in all three metrics were achieved when leveraging ACS.
The findings suggest that while using a base template in the CLIP model offers some benefits, it also has its limitations, particularly in terms of embedding concepts within a similar CLIP feature space. 
In contrast, ACS appears to reduce the uniformity between concepts, allowing the concept discovery module to discover more distinct and accurate concepts for each neuron representation.
This highlights the effectiveness of ACS in enhancing the overall performance of the concept discovery module.
\noindent\textbf{Ablation study on Concept Sensitivity ($\alpha$).}
The ablation study depicted on the left side of Figure~\ref{fig:graph} examined the impact on the F1-score by varying levels of major concept sensitivity (denoted as $\alpha$). 
We found that as concept sensitivity decreases, the F1-score initially increases($\alpha >0.95$) and then decreases. 
This pattern is due to a trade-off. Higher concept sensitivity leads to more precise concept identification but at the cost of identifying fewer concepts. 
Conversely, lower sensitivity results in more concepts being identified that may be less similar to the target concept. 
The point at which the F1-score is maximized can be seen as the optimal balance in this trade-off, providing a guideline for setting the most effective level of concept sensitivity for the major concept discovery.

\begin{figure}[t]
    \centering
    \includegraphics[width=0.48\textwidth]{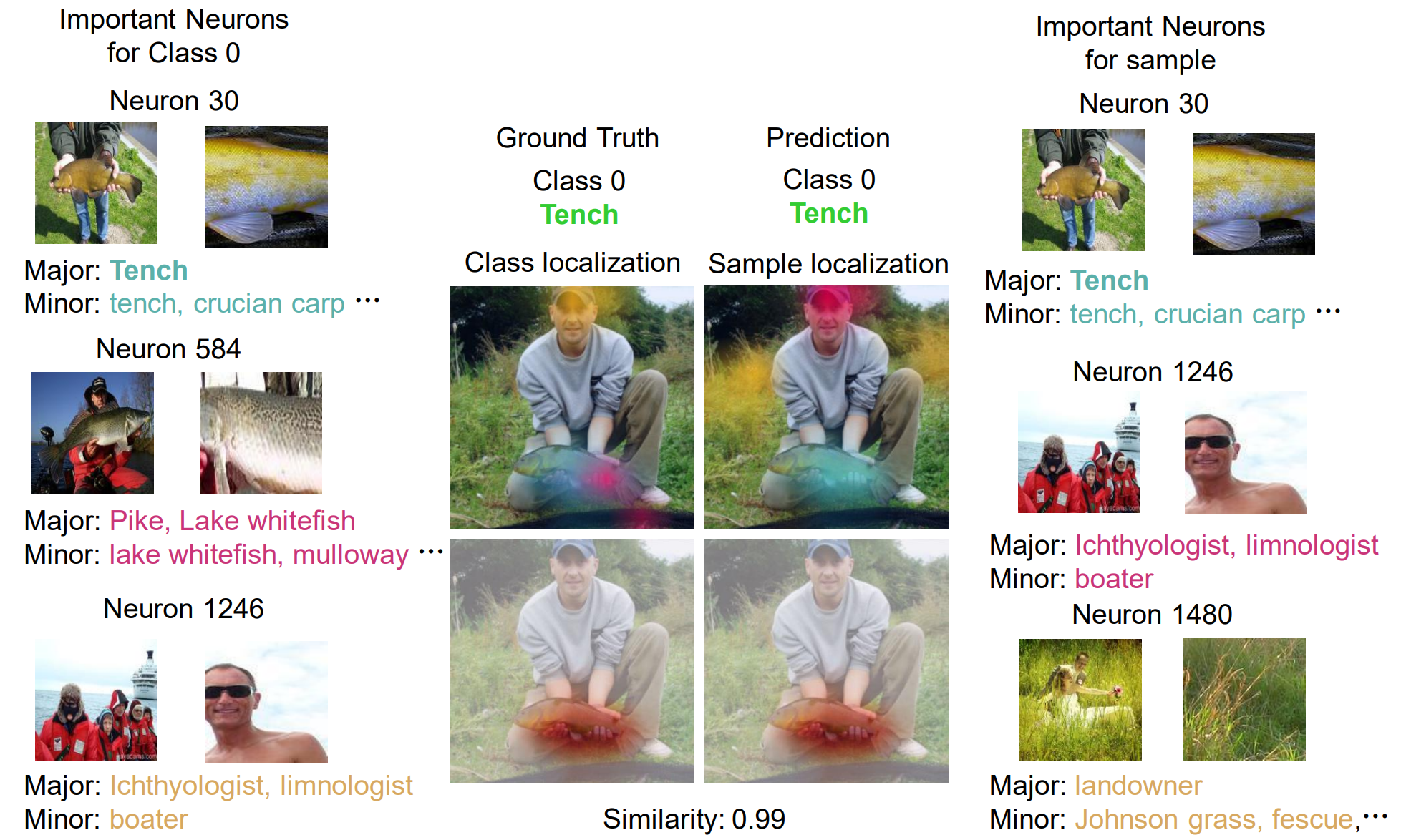}
    \caption{\small\textbf{Example of generated explanation by WWW.} From top to bottom, important neurons are displayed in the order of importance. Images in each neuron are examples of a major and minor concept, respectively. Colors in the top localization image show highly related regions for each concept. The bottom localization image is a weighted sum of important neuron activation maps displayed as a heatmap.
    }
    \label{fig:out_success} 
\end{figure}

\subsection{Discussion}
\label{sec:Discussion}

\subsubsection{Overall Explanation Generated by WWW}
\label{sec:overall_exp}
Figure~\ref{fig:out_success} illustrates an explanation example generated by WWW, where the left side of the figure provides a class explanation for a test sample by highlighting important neurons for the predicted class, and the right side depicts a sample explanation showing the important neurons identified for the test sample image prediction. 
This explanation target neurons from the penultimate layer of a ResNet-50 model pre-trained on ImageNet, using the ImageNet validation set ($D_{probe}$) and Wordnet nouns ($D_{concept}$) for the explanation. 
Notably, neuron 30 is consistently identified as the most crucial in both explanations. It represents the `tench' concept, which matches the prediction and the test sample's ground truth. 
Interestingly, while neuron 1246 is the third most important for the concept, it is the second most significant for the test sample's prediction, and neuron 1480, not highlighted in the class explanation, emerges as the third important neuron in the sample explanation.
Despite these neuron selection and ranking variations, the model accurately predicts the correct class.
The overall heatmap of the class and sample explanations can explain this. 
The heatmap explanation reveals a high cosine similarity between the two heatmaps, indicating that the calculated weighted sum of NAMs is similar.
Despite the difference in individual neuron importance, the highly related input region is remarkably similar in both cases.
\subsubsection{Anlaysis of Explanation of Failure Case}
\label{sec:analysis_exp}

\begin{figure}[t]
    \centering
    \includegraphics[width=0.46\textwidth]{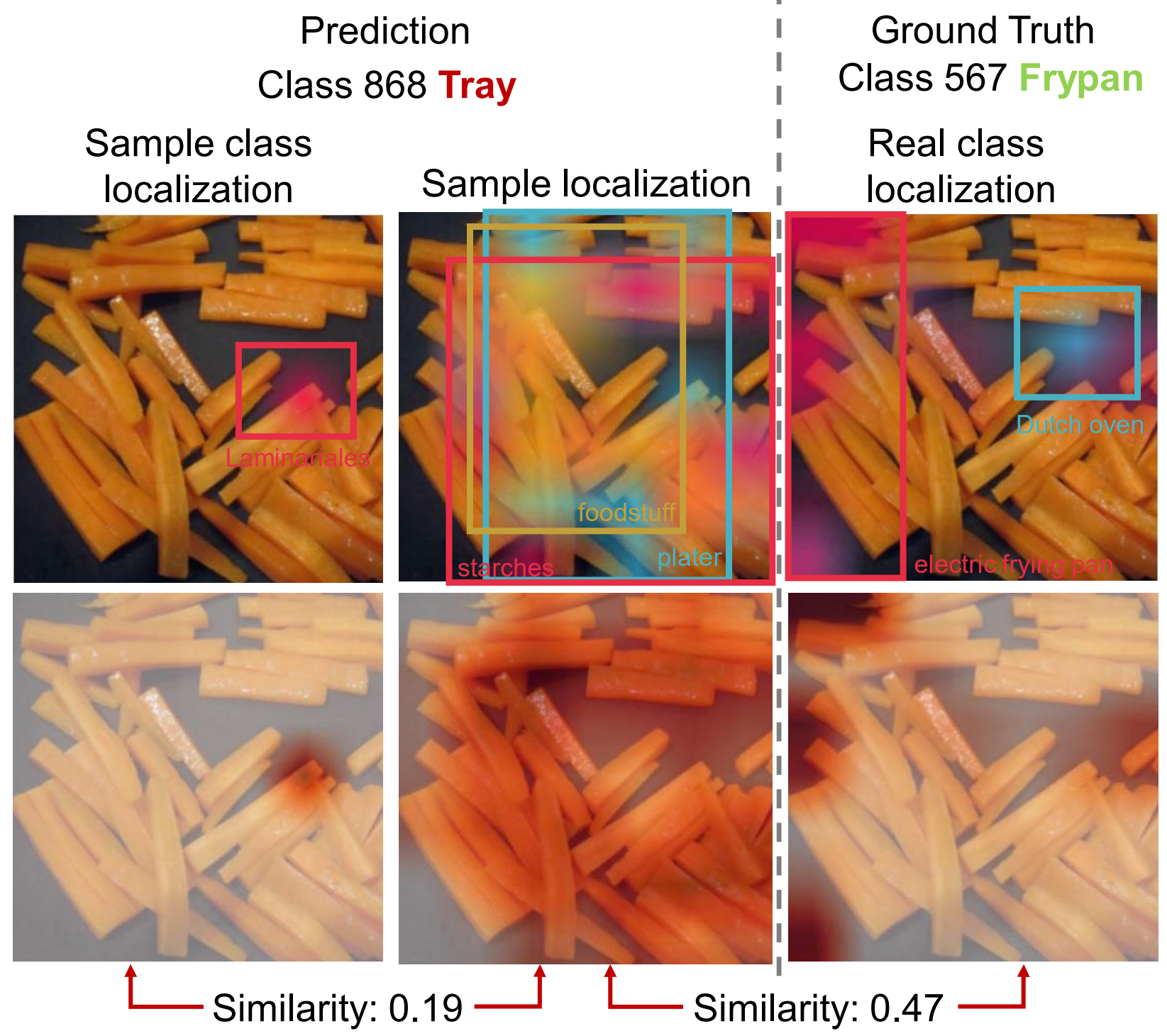}
    \caption{\small\textbf{Example of failure case explanation by WWW.} The explanations of the predicted label are presented on the left side. On the right side, the explanation of the ground-truth label is shown. In the upper half, we displayed concept attribution maps, which show important concepts and their respective regions. In the bottom half, we showed the overall heatmap which shows important regions for the model decision.
    }
    \label{fig:out_failure} 
\end{figure}

Figure~\ref{fig:out_failure} provides an example of a mispredicted sample explanation generated by the WWW. 
In this particular case of failure, not only do the selected important neurons differ between the class and sample explanations, but the cosine similarity between their respective heatmaps is relatively low, measuring at $0.19$.
Interestingly, even though the ground-truth class explanations and the sample explanations highlight different important neurons, they both localize to similar regions in the heatmap, showing a relatively high similarity score of $0.47$. 
From this observation, we explore the potential utility of heatmap similarity to predict uncertainty.

On the right side of Figure~\ref{fig:graph}, we present the results of a rejection test that evaluates heatmap similarity and the maximum softmax probability (MSP). 
When the rejection method detects the mispredicted sample correctly, we consider that as a hit. 
The number of hit is measured with respect to the rejected sample rate. 
As the rejection rate increases based on their uncertainty level, it becomes evident that heatmap similarity outperforms MSP regarding the detection of misprediction samples.
This suggests that heatmap similarity can serve as a more effective measure of uncertainty compared to MSP. 
These findings indicate that heatmap similarity can be used as a tool for predicting uncertainty.

\section{Conclusion}
We proposed WWW, a unified framework that provides comprehensive explanations for the `what', `where', and `why' of neural network decisions.
WWW demonstrates superior performance in both quantitative and qualitative measures, offering a deeper and more detailed understanding of neural network behavior. 
This is achieved through a novel integration of adaptive selection for concept discovery, neuron activation maps, and Shapley values.
WWW's adaptability is also shown across various neural network architectures, including convolutional networks and attention-based Vision Transformers. 
Additionally, our approach to predicting uncertainty through heatmap similarity analysis introduces a new way to obtain the reliability of their predictions.
By offering localized explanations with concept annotations for individual inputs, WWW enhances the transparency of the model's decision-making process, contributing to the broader goal of making AI more reliable and trustworthy. 

\section*{Acknowledgements}
This work was supported in part by the National Research Foundation of Korea (NRF) Grant funded by the Korea Government (MSIT) under Grant 2021R1G1A1094990, by the Institute of Information and Communications Technology Planning and Evaluation (IITP) Grant funded by the Korea Government (MSIT) under Grant 2022-0-00078 (Explainable Logical Reasoning for Medical Knowledge Generation), Grant IITP-2024-RS-2023-00258649 (ITRC(Information Technology Research Center) support program), Grant 2021-0-02068 (Artificial Intelligence Innovation Hub), Grant RS-2022-00155911 (Artificial Intelligence Convergence Innovation Human Resources Development (Kyung Hee University)), and by Center for Applied Research in Artificial Intelligence (CARAI) grant funded by DAPA and ADD (UD230017TD).

{
    \small
    \bibliographystyle{ieeenat_fullname}
    \bibliography{main}

\begin{thebibliography}{29}
\providecommand{\natexlab}[1]{#1}
\providecommand{\url}[1]{\texttt{#1}}
\expandafter\ifx\csname urlstyle\endcsname\relax
  \providecommand{\doi}[1]{doi: #1}\else
  \providecommand{\doi}{doi: \begingroup \urlstyle{rm}\Url}\fi

\bibitem[Achtibat et~al.(2023)Achtibat, Dreyer, Eisenbraun, Bosse, Wiegand, Samek, and Lapuschkin]{achtibat2023attribution}
Reduan Achtibat, Maximilian Dreyer, Ilona Eisenbraun, Sebastian Bosse, Thomas Wiegand, Wojciech Samek, and Sebastian Lapuschkin.
\newblock From attribution maps to human-understandable explanations through concept relevance propagation.
\newblock \emph{Nature Machine Intelligence}, 5\penalty0 (9):\penalty0 1006--1019, 2023.

\bibitem[Ahn et~al.(2023)Ahn, Park, and Kim]{ahn2023line}
Yong~Hyun Ahn, Gyeong-Moon Park, and Seong~Tae Kim.
\newblock Line: Out-of-distribution detection by leveraging important neurons.
\newblock In \emph{Proceedings of the IEEE/CVF Conference on Computer Vision and Pattern Recognition}, pages 19852--19862, 2023.

\bibitem[Bau et~al.(2017)Bau, Zhou, Khosla, Oliva, and Torralba]{bau2017network}
David Bau, Bolei Zhou, Aditya Khosla, Aude Oliva, and Antonio Torralba.
\newblock Network dissection: Quantifying interpretability of deep visual representations.
\newblock In \emph{Proceedings of the IEEE conference on computer vision and pattern recognition}, pages 6541--6549, 2017.

\bibitem[Doshi-Velez and Kim(2017)]{doshi2017towards}
Finale Doshi-Velez and Been Kim.
\newblock Towards a rigorous science of interpretable machine learning.
\newblock \emph{arXiv preprint arXiv:1702.08608}, 2017.

\bibitem[Dosovitskiy et~al.(2020)Dosovitskiy, Beyer, Kolesnikov, Weissenborn, Zhai, Unterthiner, Dehghani, Minderer, Heigold, Gelly, et~al.]{vit}
Alexey Dosovitskiy, Lucas Beyer, Alexander Kolesnikov, Dirk Weissenborn, Xiaohua Zhai, Thomas Unterthiner, Mostafa Dehghani, Matthias Minderer, Georg Heigold, Sylvain Gelly, et~al.
\newblock An image is worth 16x16 words: Transformers for image recognition at scale.
\newblock \emph{arXiv preprint arXiv:2010.11929}, 2020.

\bibitem[Fel et~al.(2023{\natexlab{a}})Fel, Boutin, Moayeri, Cadène, Bethune, andéol, Chalvidal, and Serre]{fel2023holistic}
Thomas Fel, Victor Boutin, Mazda Moayeri, Rémi Cadène, Louis Bethune, Léo andéol, Mathieu Chalvidal, and Thomas Serre.
\newblock A holistic approach to unifying automatic concept extraction and concept importance estimation.
\newblock \emph{arXiv preprint arXiv:2306.07304}, 2023{\natexlab{a}}.

\bibitem[Fel et~al.(2023{\natexlab{b}})Fel, Picard, Bethune, Boissin, Vigouroux, Colin, Cad{\`e}ne, and Serre]{fel2023craft}
Thomas Fel, Agustin Picard, Louis Bethune, Thibaut Boissin, David Vigouroux, Julien Colin, R{\'e}mi Cad{\`e}ne, and Thomas Serre.
\newblock Craft: Concept recursive activation factorization for explainability.
\newblock In \emph{Proceedings of the IEEE/CVF Conference on Computer Vision and Pattern Recognition}, pages 2711--2721, 2023{\natexlab{b}}.

\bibitem[Fellbaum(2005)]{wordnet}
Christiane Fellbaum.
\newblock Wordnet and wordnets. encyclopedia of language and linguistics, 2005.

\bibitem[Fong and Vedaldi(2018)]{fong2018net2vec}
Ruth Fong and Andrea Vedaldi.
\newblock Net2vec: Quantifying and explaining how concepts are encoded by filters in deep neural networks.
\newblock In \emph{Proceedings of the IEEE conference on computer vision and pattern recognition}, pages 8730--8738, 2018.

\bibitem[Fong and Vedaldi(2017)]{fong2017interpretable}
Ruth~C Fong and Andrea Vedaldi.
\newblock Interpretable explanations of black boxes by meaningful perturbation.
\newblock In \emph{Proceedings of the IEEE international conference on computer vision}, pages 3429--3437, 2017.

\bibitem[He et~al.(2016)He, Zhang, Ren, and Sun]{resnet}
Kaiming He, Xiangyu Zhang, Shaoqing Ren, and Jian Sun.
\newblock Identity mappings in deep residual networks.
\newblock In \emph{Proceedings of the European conference on computer vision}, pages 630--645. Springer, 2016.

\bibitem[Hernandez et~al.(2021)Hernandez, Schwettmann, Bau, Bagashvili, Torralba, and Andreas]{MILAN}
Evan Hernandez, Sarah Schwettmann, David Bau, Teona Bagashvili, Antonio Torralba, and Jacob Andreas.
\newblock Natural language descriptions of deep visual features.
\newblock In \emph{International Conference on Learning Representations}, 2021.

\bibitem[Jacovi et~al.(2021)Jacovi, Marasovi{\'c}, Miller, and Goldberg]{jacovi2021formalizing}
Alon Jacovi, Ana Marasovi{\'c}, Tim Miller, and Yoav Goldberg.
\newblock Formalizing trust in artificial intelligence: Prerequisites, causes and goals of human trust in ai.
\newblock In \emph{Proceedings of the 2021 ACM conference on fairness, accountability, and transparency}, pages 624--635, 2021.

\bibitem[Kalibhat et~al.(2023)Kalibhat, Bhardwaj, Bruss, Firooz, Sanjabi, and Feizi]{kalibhat2023identifying}
Neha Kalibhat, Shweta Bhardwaj, C~Bayan Bruss, Hamed Firooz, Maziar Sanjabi, and Soheil Feizi.
\newblock Identifying interpretable subspaces in image representations.
\newblock \emph{International Conference on Machine Learning}, 2023.

\bibitem[Kaminski and Urban(2021)]{kaminski2021right}
Margot~E Kaminski and Jennifer~M Urban.
\newblock The right to contest ai.
\newblock \emph{Columbia Law Review}, 121\penalty0 (7):\penalty0 1957--2048, 2021.

\bibitem[Khakzar et~al.(2021)Khakzar, Baselizadeh, Khanduja, Rupprecht, Kim, and Navab]{khakzar2021neural}
Ashkan Khakzar, Soroosh Baselizadeh, Saurabh Khanduja, Christian Rupprecht, Seong~Tae Kim, and Nassir Navab.
\newblock Neural response interpretation through the lens of critical pathways.
\newblock In \emph{Proceedings of the IEEE/CVF Conference on Computer Vision and Pattern Recognition}, pages 13528--13538, 2021.

\bibitem[Kim et~al.(2018)Kim, Wattenberg, Gilmer, Cai, Wexler, Viegas, et~al.]{kim2018interpretability}
Been Kim, Martin Wattenberg, Justin Gilmer, Carrie Cai, James Wexler, Fernanda Viegas, et~al.
\newblock Interpretability beyond feature attribution: Quantitative testing with concept activation vectors (tcav).
\newblock In \emph{International conference on machine learning}, pages 2668--2677. PMLR, 2018.

\bibitem[Kop(2021)]{kop2021eu}
Mauritz Kop.
\newblock Eu artificial intelligence act: The european approach to ai.
\newblock Stanford-Vienna Transatlantic Technology Law Forum, Transatlantic Antitrust~…, 2021.

\bibitem[Mu and Andreas(2020)]{mu2020compositional}
Jesse Mu and Jacob Andreas.
\newblock Compositional explanations of neurons.
\newblock \emph{Advances in Neural Information Processing Systems}, 33:\penalty0 17153--17163, 2020.

\bibitem[Oikarinen and Weng(2023)]{oikarinen2023clipdissect}
Tuomas Oikarinen and Tsui-Wei Weng.
\newblock {CLIP}-dissect: Automatic description of neuron representations in deep vision networks.
\newblock In \emph{The Eleventh International Conference on Learning Representations}, 2023.

\bibitem[Radford et~al.(2021)Radford, Kim, Hallacy, Ramesh, Goh, Agarwal, Sastry, Askell, Mishkin, Clark, et~al.]{radford2021clip}
Alec Radford, Jong~Wook Kim, Chris Hallacy, Aditya Ramesh, Gabriel Goh, Sandhini Agarwal, Girish Sastry, Amanda Askell, Pamela Mishkin, Jack Clark, et~al.
\newblock Learning transferable visual models from natural language supervision.
\newblock In \emph{International conference on machine learning}, pages 8748--8763. PMLR, 2021.

\bibitem[Russakovsky et~al.(2015)Russakovsky, Deng, Su, Krause, Satheesh, Ma, Huang, Karpathy, Khosla, Bernstein, et~al.]{imagenet}
Olga Russakovsky, Jia Deng, Hao Su, Jonathan Krause, Sanjeev Satheesh, Sean Ma, Zhiheng Huang, Andrej Karpathy, Aditya Khosla, Michael Bernstein, et~al.
\newblock Imagenet large scale visual recognition challenge.
\newblock \emph{International journal of computer vision}, 115:\penalty0 211--252, 2015.

\bibitem[Selvaraju et~al.(2017)Selvaraju, Cogswell, Das, Vedantam, Parikh, and Batra]{selvaraju2017grad}
Ramprasaath~R Selvaraju, Michael Cogswell, Abhishek Das, Ramakrishna Vedantam, Devi Parikh, and Dhruv Batra.
\newblock Grad-cam: Visual explanations from deep networks via gradient-based localization.
\newblock In \emph{Proceedings of the IEEE international conference on computer vision}, pages 618--626, 2017.

\bibitem[Shapley(1997)]{shapley1997value}
Lloyd~S Shapley.
\newblock A value for n-person games.
\newblock \emph{Classics in game theory}, 69, 1997.

\bibitem[Sloan(2010)]{sloan2010aristotle}
Michael~C Sloan.
\newblock Aristotle’s nicomachean ethics as the original locus for the septem circumstantiae.
\newblock \emph{Classical Philology}, 105\penalty0 (3):\penalty0 236--251, 2010.

\bibitem[Song~et al.(2020)]{song2020mpnet}
K. Song~et al.
\newblock Mpnet: Masked and permuted pre-training for language understanding.
\newblock \emph{NeurIPS}, 33:\penalty0 16857--16867, 2020.

\bibitem[Wang et~al.(2022)Wang, Lee, and Qi]{wang2022hint}
Andong Wang, Wei-Ning Lee, and Xiaojuan Qi.
\newblock Hint: Hierarchical neuron concept explainer.
\newblock In \emph{Proceedings of the IEEE/CVF Conference on Computer Vision and Pattern Recognition}, pages 10254--10264, 2022.

\bibitem[Zhang et~al.(2021{\natexlab{a}})Zhang, Khakzar, Li, Farshad, Kim, and Navab]{zhang2021fine}
Yang Zhang, Ashkan Khakzar, Yawei Li, Azade Farshad, Seong~Tae Kim, and Nassir Navab.
\newblock Fine-grained neural network explanation by identifying input features with predictive information.
\newblock \emph{Advances in Neural Information Processing Systems}, 34:\penalty0 20040--20051, 2021{\natexlab{a}}.

\bibitem[Zhang et~al.(2021{\natexlab{b}})Zhang, Ti{\v{n}}o, Leonardis, and Tang]{zhang2021survey}
Yu Zhang, Peter Ti{\v{n}}o, Ale{\v{s}} Leonardis, and Ke Tang.
\newblock A survey on neural network interpretability.
\newblock \emph{IEEE Transactions on Emerging Topics in Computational Intelligence}, 5\penalty0 (5):\penalty0 726--742, 2021{\natexlab{b}}.

\end{thebibliography}
}

\clearpage
\setcounter{page}{1}
\maketitlesupplementary

\setcounter{table}{0}
\renewcommand{\thetable}{S.\arabic{table}}

\setcounter{figure}{0}
\renewcommand{\thefigure}{S.\arabic{figure}}

\setcounter{equation}{0}
\renewcommand{\theequation}{S.\arabic{equation}}

\appendix

\section{Implementation Details}
In this section, we discuss implementation details for the experiment.
\subsection{ResNet-50 trained on ImageNet}
\label{impl_res50}
We use ResNet-50 of ImageNet-1k pre-trained weight on the torchvision 0.14.0 version. \\
\textbf{WWW (Ours).}
We select 40 high-activating images and 40 high-activating crop images for major and minor concepts for each neuron, respectively.
Adaptive threshold $\alpha$ is set to 0.95 for major concept selection and $\alpha$ is set to 0.90 for minor concept selection.
\\
\textbf{CLIP-Dissect~\cite{oikarinen2023clipdissect}.}
We implemented from the official GitHub code (\href{https://github.com/Trustworthy-ML-Lab/CLIP-dissect}{https://github.com/Trustworthy-ML-Lab/CLIP-dissect}). We only change pre-trained model weight.
\\
\textbf{MILAN(b)~\cite{MILAN}.}
We implemented from the official GitHub code (\href{https://github.com/evandez/neuron-descriptions}{https://github.com/evandez/neuron-descriptions}). We only change pre-trained model weight.
\\
\textbf{FALCON~\cite{kalibhat2023identifying}.}
When matching neuron concepts with official FALCON threshold, the concepts are not matched to all final layer neurons. So, we modified the threshold to 0.35 and implemented it to match concepts to most neurons in the ResNet-50 final layer. Additionally, to evaluate concept matching performance for each neuron, the official FALCON of matching concepts to a group of neurons was modified and implemented to match concepts only to each neuron.
\\

\subsection{ViT-B/16 trained on ImageNet}
\label{impl_vit}
We use the pre-trained weight of ViT B/16 on the ImageNet dataset in the timm 0.9.7 version. \\
\textbf{WWW (Ours).}
We select 40 high-activating images and 40 high-activating crop images for major and minor concepts for each neuron, respectively.
Adaptive threshold $\alpha$ is set to 0.95 for major concept selection and $\alpha$ is set to 0.90 for minor concept selection.
\\
\textbf{CLIP-Dissect~\cite{oikarinen2023clipdissect}.}
We implemented from the official GitHub code (\href{https://github.com/Trustworthy-ML-Lab/CLIP-dissect}{https://github.com/Trustworthy-ML-Lab/CLIP-dissect}). We only change pre-trained model weight.
\\

\subsection{ResNet-18 trained on Places365}
\label{impl_res18}
We used the same ResNet-18 pre-trained in places365 dataset weight used in CLIP-Dissect official implementation
(\href{https://github.com/Trustworthy-ML-Lab/CLIP-dissect}{https://github.com/Trustworthy-ML-Lab/CLIP-dissect}).\\
\textbf{WWW (Ours).}
We select 40 high-activating images and 40 high-activating crop images for major and minor concepts for each neuron, respectively.
Adaptive threshold $\alpha$ is set to 0.95 for major concept selection and $\alpha$ is set to 0.90 for minor concept selection.
\\
\textbf{CLIP-Dissect~\cite{oikarinen2023clipdissect}.}
We implemented from the official GitHub code (\href{https://github.com/Trustworthy-ML-Lab/CLIP-dissect}{https://github.com/Trustworthy-ML-Lab/CLIP-dissect}).

\section{Ablation Studies}
\subsection{Ablation study on the Number of High Activating Images}
\begin{figure}[t]
    \centering
    \includegraphics[width=0.4\textwidth]{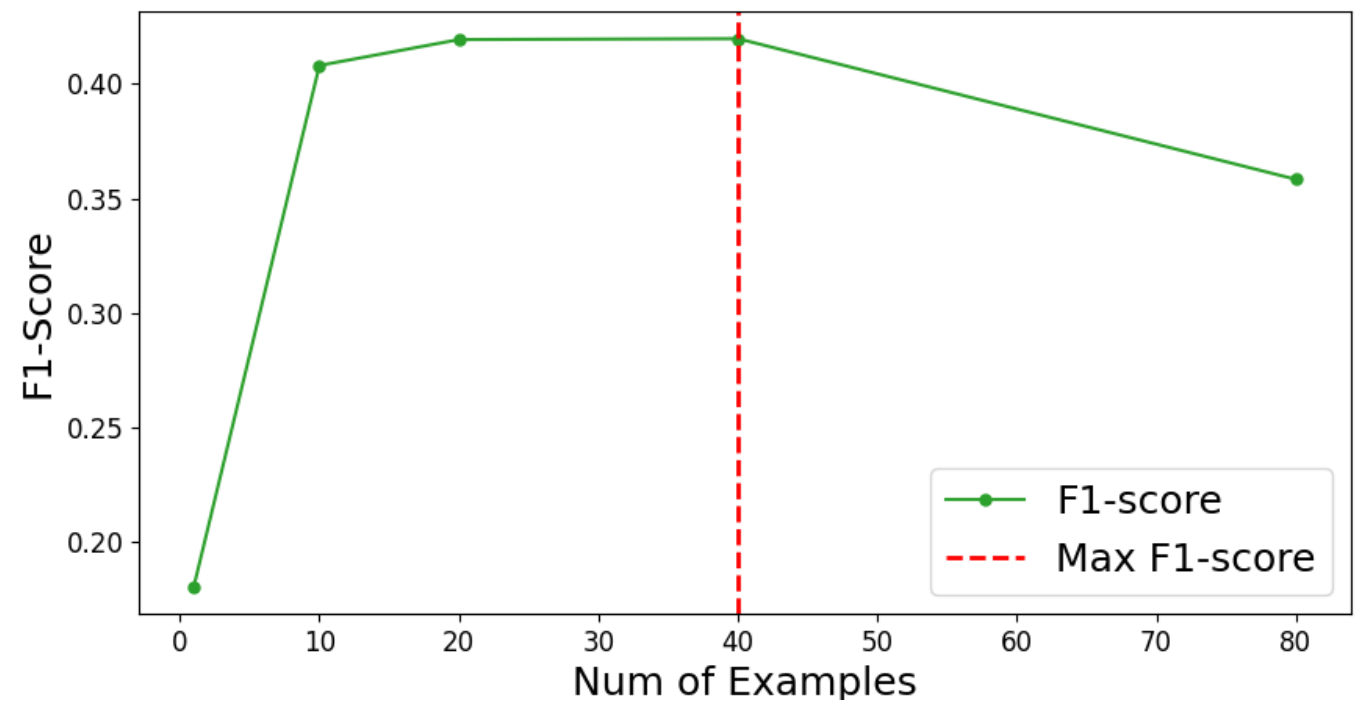}
    \caption{\small\textbf{Ablation study on the number of selected examples.} We displayed F1-score change regarding the number of selected example images for each neuron. The red line refers to the point that maximizes F1 score. 
    }
    \label{fig:num_graph} 
\end{figure}

In this section, we are going to discuss the effect of changing the numbers of selected example-based representations(i.e., high-activating samples).\\
\textbf{Implementation Details}.
We used ImageNet-1k pre-trained ResNet-50, ImageNet Validation set as $D_{probe}$, and Wordnet nouns as $D_{concept}$.
For hyperparameter settings, We used the same settings in~\ref{impl_res50}\\
\textbf{Results}. In figure~\ref{fig:num_graph}, we show the relation between the number of selected images and performance.
With the small number of examples (i.e., $k = 1$), WWW shows low performance.
That is because, in one single example image, there are dozens of different concepts which is not related to the neuron representation. 
But with a sufficient number of examples, the F1-score showed the best performance.
At a large number of examples (i.e., $k = 80$), WWW showed decreased performance due to the less similar concept images.

\subsection{Ablation Study on the ACS to Other Baselines}
In this section, we compared four different methods for selecting concepts: Cosine similarity, L1, L2, and WWW (ACS + AT).
Cosine similarity refers to the concept that has the highest cosine similarity with images selected as a concept.
L1 and L2 refer to selecting a concept that has minimal L1 and L2 errors with images, respectively.
For WWW, we used Adaptive Cosine Similarity (ACS) and Adaptive Thresholding (AT) for concept selection.\\
\textbf{Implementation Details}.
We used ImageNet-1k pre-trained ResNet-50, ImageNet Validation set as $D_{probe}$, and Wordnet nouns as $D_{concept}$.
For hyperparameter settings, We used the same settings in~\ref{impl_res50}\\
\textbf{Results}.
In table~\ref{tab:Abl_cs}, WWW shows the best performance, except for the Hit rate, compared to other baselines.

\begin{table}[t]
\caption{\textbf{Ablation on concept selection methods.}
We compared Cosine similarity, L1, L2, and WWW (ACS + AT) and their performance on four different metrics. We used ResNet-50, ImageNet validation($D_{probe}$), Wordnet nouns($D_{concept}$). 
\textbf{Bold} numbers represent the best scores between the same settings. 
The average score and standard errors of the $1000$ final layer neurons are reported.
}
\centering
\scalebox{0.75}{
\begin{tabular}{@{}l|ccc}
\toprule
Method & CLIP cos & mpnet cos & F1-score \\
\midrule
Cosine & $0.8499\pm{0.003}$ & $0.6123\pm{0.007}$ & $0.3265\pm{0.009}$ \\  
L1     & $0.8497\pm{0.003}$ & $0.6033\pm{0.008}$ & $0.2644\pm{0.013}$ \\
L2     & $0.6331\pm{0.006}$ & $0.3861\pm{0.008}$ & $0.1112\pm{0.009}$ \\ 
WWW (Ours) & $\mathbf{0.8858\pm{0.003}}$ & $\mathbf{0.6945\pm{0.008}}$ & $\mathbf{0.4197\pm{0.012}}$\\
\bottomrule
\end{tabular}
}
\label{tab:Abl_cs}
\end{table}

\section{Explanation Case Analysis}
\begin{figure*}[t]
    \centering
    \includegraphics[width=0.8\textwidth]{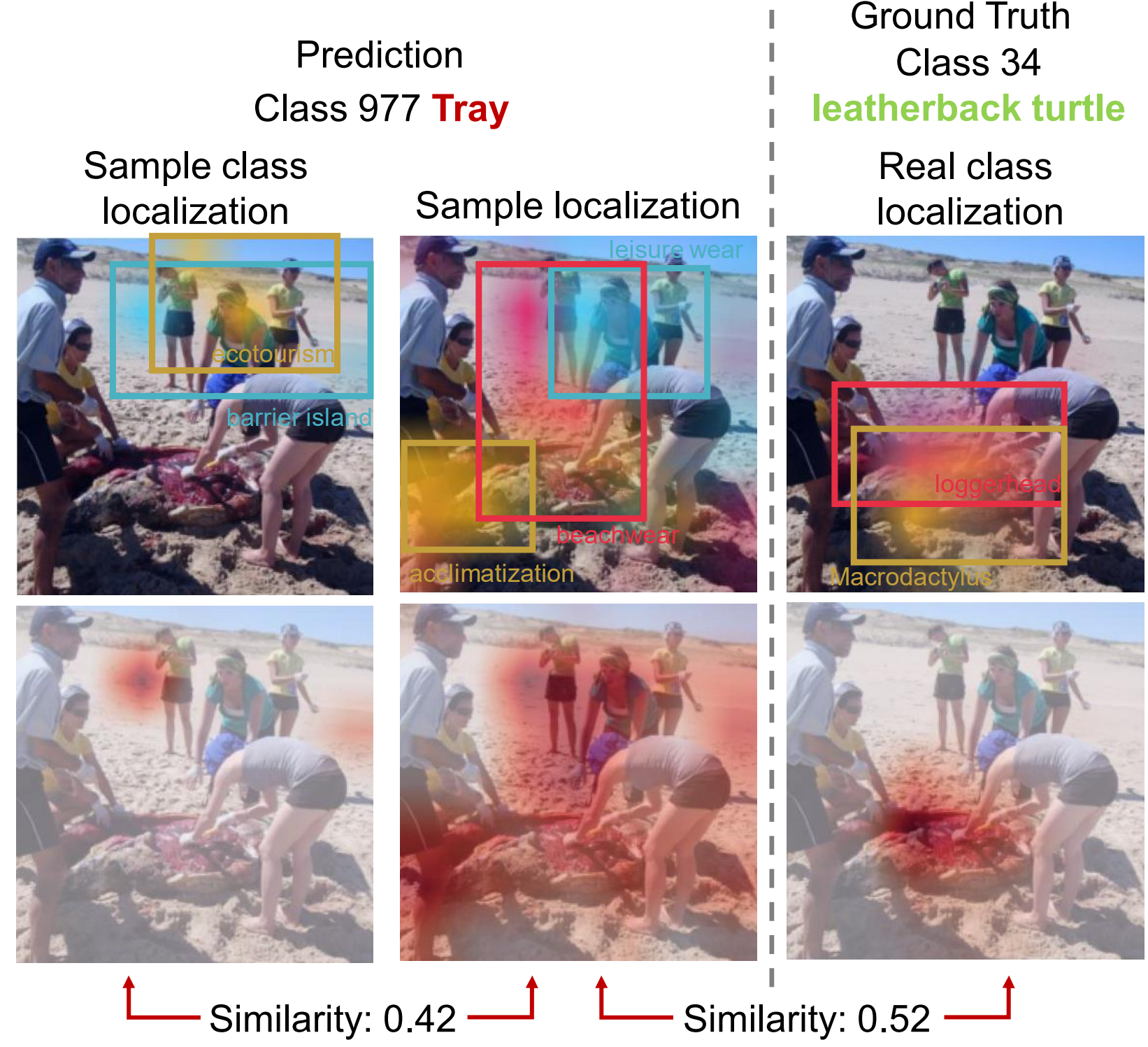}
    \caption{\small\textbf{Example of failure case explanation by WWW.} The explanations of the predicted label are presented on the left side. On the right side, the explanation of the ground-truth label is shown. We displayed related regions of the concept as bounding boxes in the order of blue, red, and yellow boxes. (blue represents the most important concept) At the bottom, we showed the similarity between the two heatmaps.
    }
    \label{fig:out_failure_s1} 
\end{figure*}

\begin{figure*}[t]
    \centering
    \includegraphics[width=0.8\textwidth]{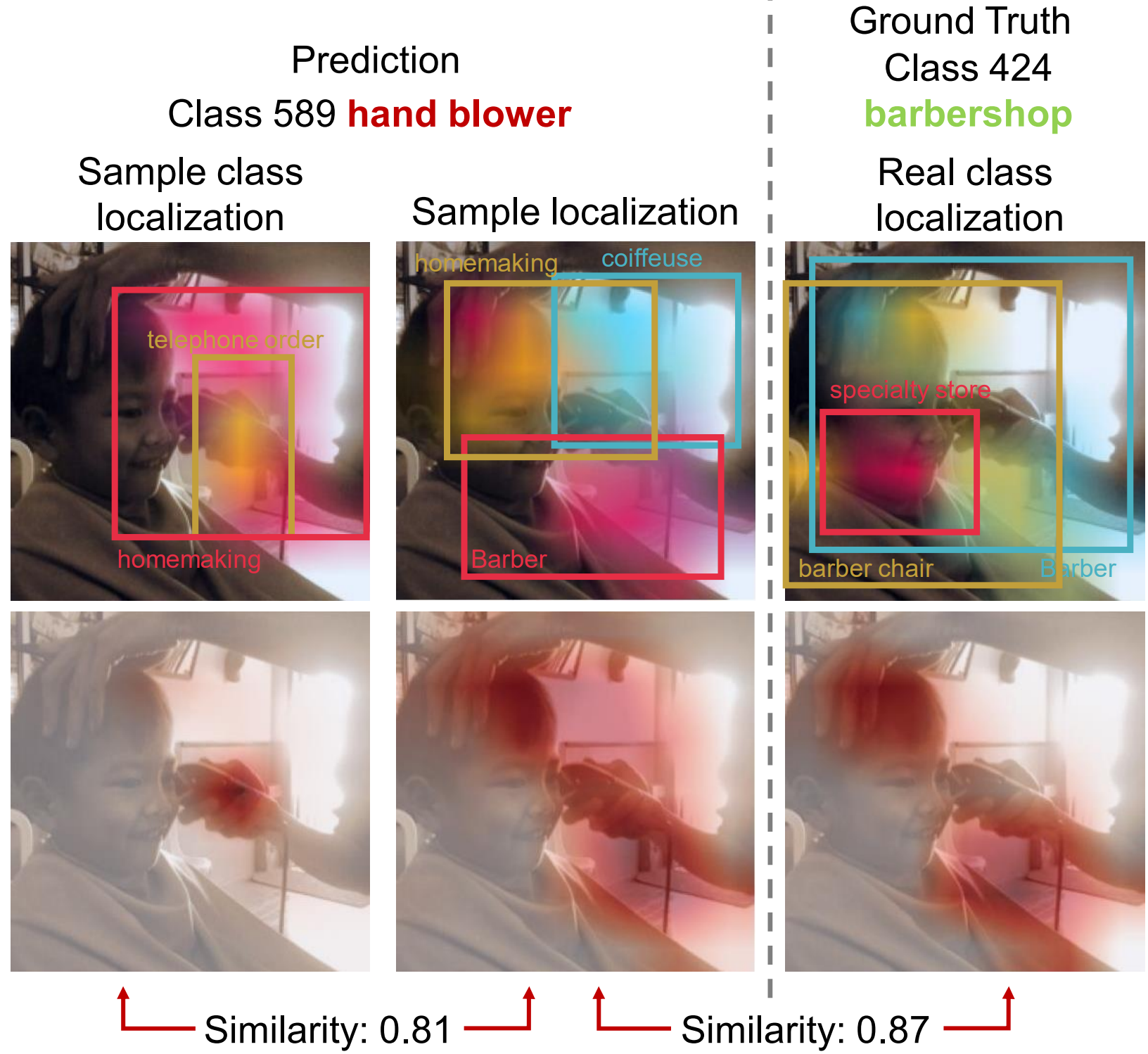}
    \caption{\small\textbf{Example of failure case explanation by WWW.} The explanations of the predicted label are presented on the left side. On the right side, the explanation of the ground-truth label is shown. We displayed related regions of the concept as bounding boxes in the order of blue, red, and yellow boxes. (blue represents the most important concept) At the bottom, we showed the similarity between the two heatmaps.
    }
    \label{fig:out_failure_s2} 
\end{figure*}

In this section, we displayed sample cases for explanation generated by WWW.
We showed two additional failure cases in figure~\ref{fig:out_failure_s1} and figure~\ref{fig:out_failure_s2}.
In both cases, not only do the selected important neurons differ between the class and sample explanations, but the cosine similarity between their respective heatmaps is relatively low.
Even though the ground-truth class explanations and the sample explanations highlight different important neurons, they both localize to similar regions in the heatmap, showing a relatively high similarity score in both cases.
\\
\textbf{Implemenetation Details.}
We used ImageNet-1k pre-trained ResNet-50, ImageNet Validation set as $D_{probe}$, and Wordnet nouns as $D_{concept}$.
For hyperparameter settings, We used the same settings in~\ref{impl_res50}

\section{Qualitative Results}
In section~\ref{Sec:Qual_res50}, we displayed concept selection results on different layers of ResNet-50. 
In section~\ref{Sec:Qual_vit}, we displayed concept selection results on the final layer of ViT/B-16 model.\\
\textbf{Implemenetation Details.}
For the ResNet-50 qualitative result, we used the same settings as Section~\ref{impl_res50}.
We used ImageNet-1k pre-trained ResNet-50, ImageNet Validation set as $D_{probe}$, and Wordnet nouns as $D_{concept}$.
For the ViT/b-16 qualitative result, we used the same settings as Section~\ref{impl_vit}.
We also used the ImageNet pre-trained ViT/B-16 model, ImageNet Validation set as $D_{probe}$, and Wordnet nouns as $D_{concept}$.
\subsection{ResNet-50}
\label{Sec:Qual_res50}
\begin{figure*}[t!]
    \centering
    \includegraphics[width=\textwidth]{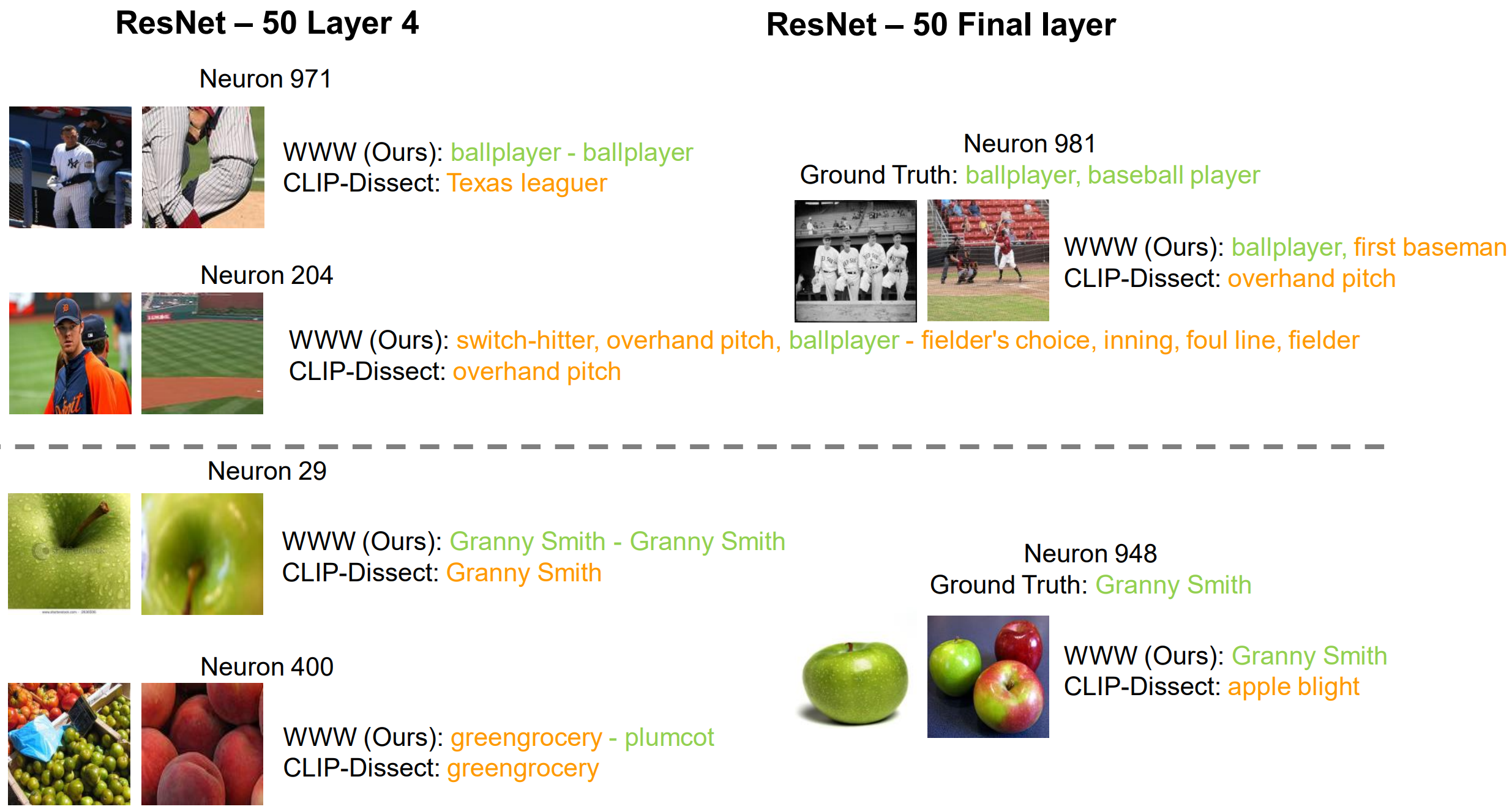}
    \caption{\small\textbf{Qualitative comparison of WWW.} We compared WWW with competitive baseline (CLIP-Dissection\cite{oikarinen2023clipdissect}, in two final layer neurons and four penultimate layer (i.e., layer 4) neurons with each neuron's highly activating images. Penultimate layer neurons are top-$2$ important neurons of the final layer class. We have colored the descriptions green if they match the images, yellow if they match but are too generic or similar, and red if they do not match.
    }
    \label{fig:qual_s1} 
\end{figure*}

\begin{figure*}[t!]
    \centering
    \includegraphics[width=\textwidth]{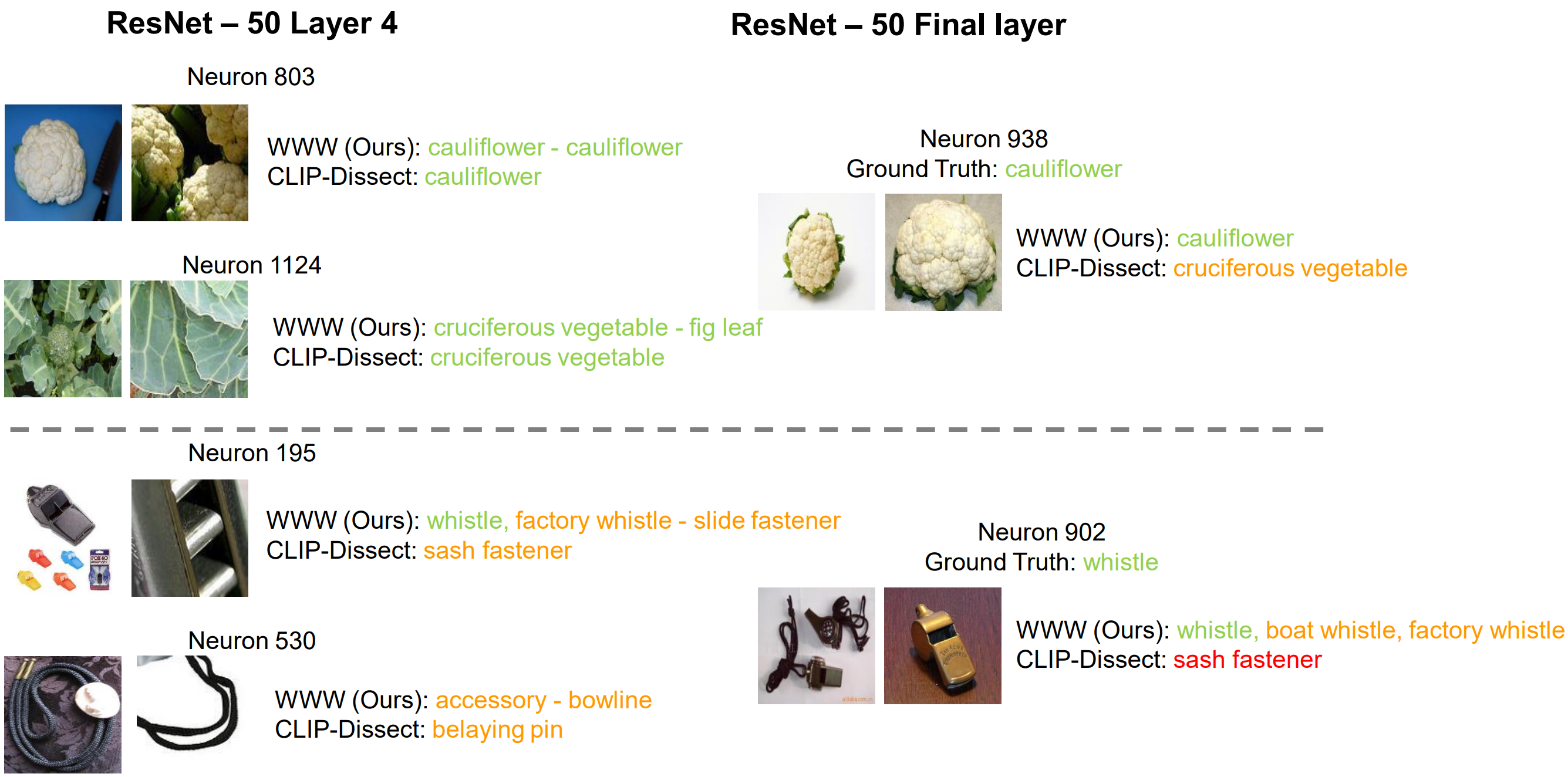}
    \caption{\small\textbf{Qualitative comparison of WWW.} We compared WWW with competitive baseline (CLIP-Dissection\cite{oikarinen2023clipdissect}, in two final layer neurons and four penultimate layer (i.e., layer 4) neurons with each neuron's highly activating images. Penultimate layer neurons are top-$2$ important neurons of the final layer class. We have colored the descriptions green if they match the images, yellow if they match but are too generic or similar, and red if they do not match.
    }
    \label{fig:qual_s2} 
\end{figure*}

In this section, we displayed the qualitative results of WWW on ResNet-50.
In figure~\ref{fig:qual_s1} and figure~\ref{fig:qual_s2} shows examples of descriptions for hidden neurons in the penultimate and final layers.
Neurons in the penultimate layer are top-$2$ important neurons of the final layer neuron's ground truth label class.
We observed that WWW not only interpreted each neuron well but also showed robust interpretation that the most important neuron of the class in the penultimate layer represents the exact same major concept as the final layer neuron.


\begin{figure*}[t]
    \centering
    \includegraphics[width=1.0\textwidth]{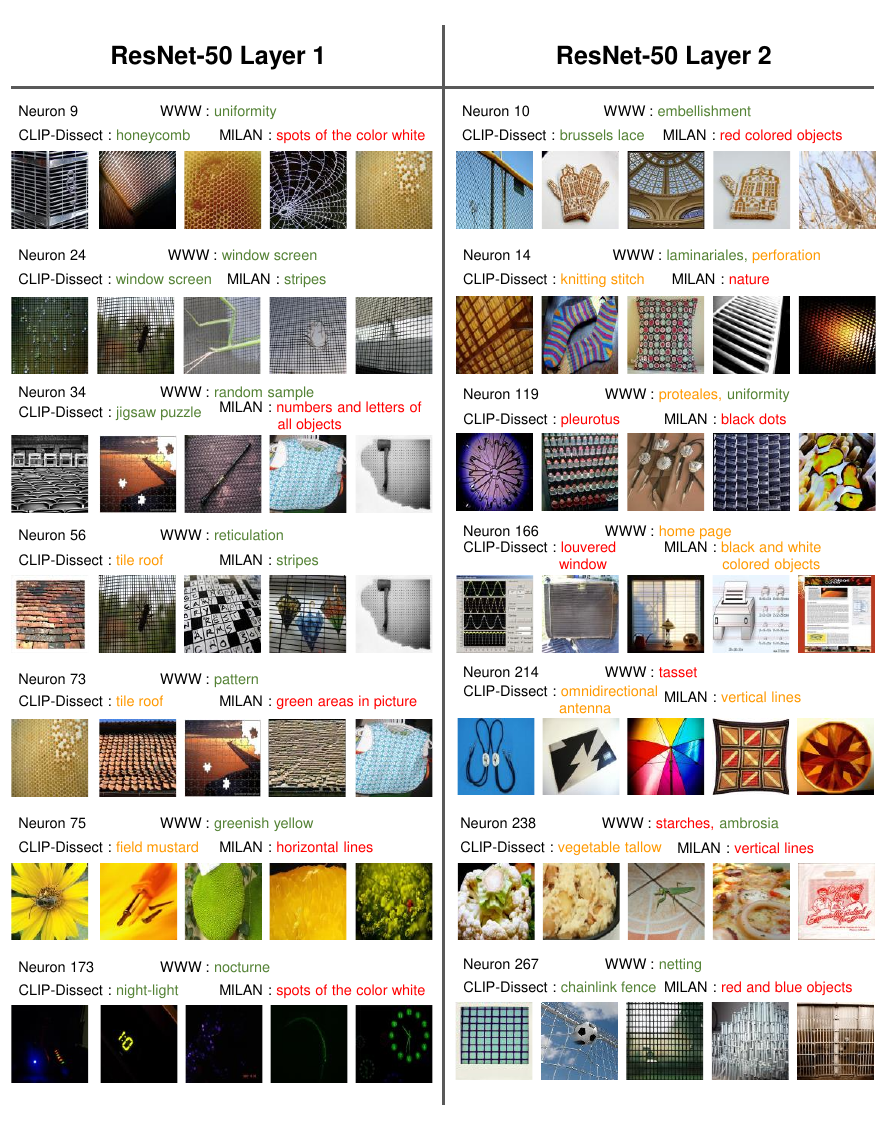}
    \caption{\small\textbf{Qualitative results on the first and second layer of ResNet-50.} We displayed each neuron's five highest activating examples with the ground truth label. We have colored the descriptions green if they match the images, yellow if they match but are too generic or similar, and red if they do not match.
    }
    \label{fig:WWW_Res_12} 
\end{figure*}

\begin{figure*}[t]
    \centering
    \includegraphics[width=1.0\textwidth]{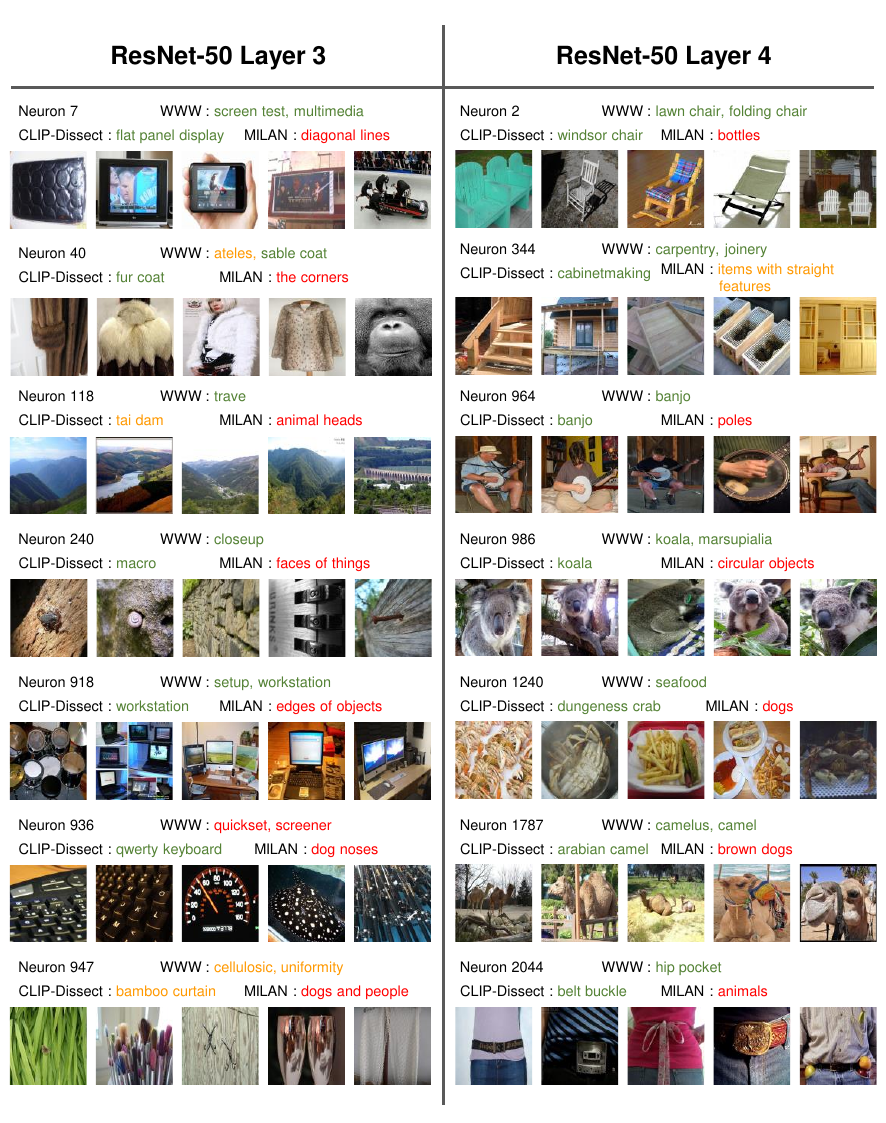}
    \caption{\small\textbf{Qualitative results on the third and fourth layer of ResNet-50.} We displayed each neuron's five highest activating examples with the ground truth label. We have colored the descriptions green if they match the images, yellow if they match but are too generic or similar, and red if they do not match.
    }
    \label{fig:WWW_Res_34} 
\end{figure*}

In figure~\ref{fig:WWW_Res_12} and figure~\ref{fig:WWW_Res_34}, we showed each neuron's 5 highest activating examples with the ground truth label. 
We have colored the descriptions green if they match the images, yellow if they match but are too generic or similar, and red if they do not match.

\subsection{ViT/b-16}
\label{Sec:Qual_vit}
\begin{figure*}[t]
    \centering
    \includegraphics[width=1.0\textwidth]{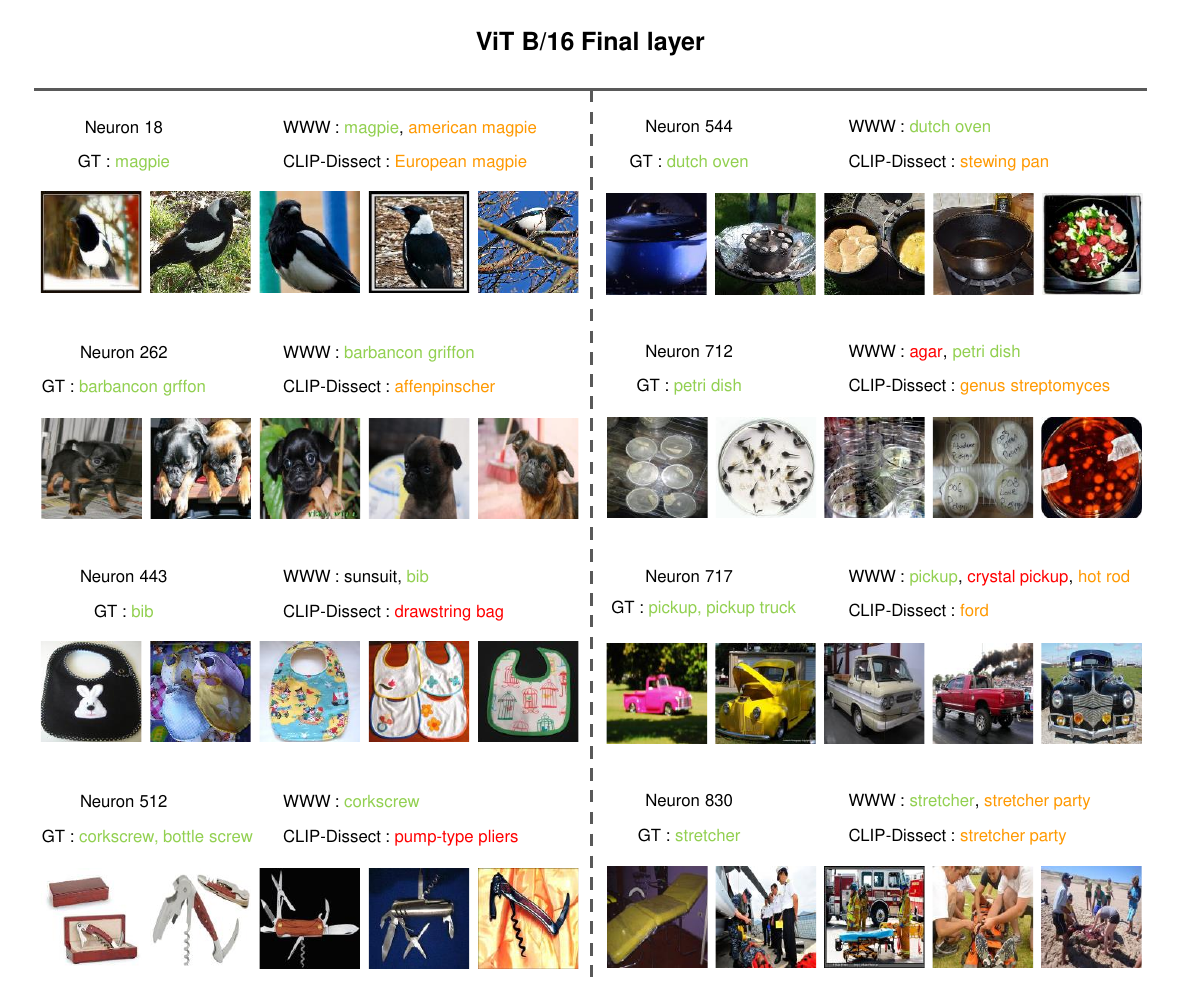}
    \caption{\small\textbf{Qualitative results on the final layer of ViT/B-16.} We displayed each neuron's five highest activating examples with the ground truth label. We have colored the descriptions green if they match the images, yellow if they match but are too generic or similar, and red if they do not match.
    }
    \label{fig:WWW_ViT_supple} 
\end{figure*}

In this section, we displayed the qualitative results of WWW on Vision transformers.
We showed each neuron's five highest activating examples with the ground truth label. 
We have colored the descriptions green if they match the images, yellow if they match but are too generic or similar, and red if they do not match.
In figure~\ref{fig:WWW_ViT_supple}, WWW exactly matches the ground truth label, outperforming the other baseline.

\section{t-tests in tables}
We conducted paired t-tests to measure the statistical significance of the differences.
Over Tables 2 and 3, the performance gap between WWW and the most comparable method (i.e., CLIP-Dissect) is statistically significant for both metrics of CLIP cos ($p<0.001$) and F1-score ($p<0.001$) in settings of $D_{probe} = $ ImageNet validation with $D_{concept} = $ Wordnet (80k).
In Table 4, WWW significantly improves CLIP cos compared to CLIP-Dissect in the setting of $D_{probe} = $ ImageNet validation with $D_{concept} = $ Wordnet 80k ($p=0.018$).

\section{Heatmap similarity and misprediction detection based on uncertainty measure}
\begin{table}[t!]
\caption{\textbf{Experiment on AUROC of MSP and our method for mis-prediction detection.}
}
\vspace{-0.1cm}
\centering
\scalebox{0.9}{
\begin{tabular}{lc}
\toprule
Method & AUROC\\
\midrule
MSP & 0.808 \\
\textbf{WWW (Ours)} & \textbf{0.903} \\
\bottomrule
\end{tabular}
}
\vspace{-0.1cm}
\label{tab:AUROC}
\end{table}

In misprediction detection, our approach is to detect uncertain samples (i.e., samples with low heatmap similarity), which can be a highly potential failure case.
We conducted a 'large-scale' experiment to quantify the quality of the Reasoning module.
To show the distribution of the similarities across correct predictions and mispredictions,
we calculated AUROC on both MSP and  WWW (i.e., heatmap similarity) for the binary classification task of misprediction based on estimated uncertainty.
We used the ImageNet pre-trained ResNet-50 model and ImageNet validation set.
As in Table \ref{tab:AUROC}, WWW shows outstanding performance compared to the MSP.
This result indicates that WWW can be used as a misprediction detector.

\end{document}